%% file: main.tex
\documentclass[10pt,twocolumn,letterpaper]{article}

\usepackage{iccv}
\usepackage{times}
\usepackage{epsfig}
\usepackage{graphicx}
\usepackage{amsmath}
\usepackage{amssymb}

\usepackage[accsupp]{axessibility}  
\usepackage{multirow}
\usepackage{subcaption}
\usepackage{placeins}
\usepackage[pagebackref=true,breaklinks=true,letterpaper=true,colorlinks,bookmarks=false]{hyperref}

\iccvfinalcopy 

\def\iccvPaperID{3731} 

\ificcvfinal\fi

\begin{document}

\title{SPatchGAN: A Statistical Feature Based Discriminator for Unsupervised Image-to-Image Translation}

\author{Xuning Shao, Weidong Zhang\\
NetEase Games AI Lab\\
599 Wangshang Road, Binjiang District, Hangzhou, P.R. China\\
{\tt\small \{shaoxuning, zhangweidong02\}@corp.netease.com}
}

\maketitle
\ificcvfinal\thispagestyle{empty}\fi

\input{sections/abstract}

\input{sections/introduction}
\input{sections/related_work}

\input{sections/model}

\input{sections/experiments}

\input{sections/conclusions}

\FloatBarrier

{\small
\bibliographystyle{ieee_fullname}
\bibliography{sections/references}
}

\clearpage
\FloatBarrier

\input{sections/supl}

\end{document}

%% file: sections/abstract.tex
\begin{abstract}
For unsupervised image-to-image translation, we propose a discriminator architecture which focuses on the statistical features instead of individual patches. The network is stabilized by distribution matching of key statistical features at multiple scales. Unlike the existing methods which impose more and more constraints on the generator, our method facilitates the shape deformation and enhances the fine details with a greatly simplified framework. We show that the proposed method outperforms the existing state-of-the-art models in various challenging applications including selfie-to-anime, male-to-female and glasses removal.
\end{abstract}

%% file: sections/introduction.tex
\section{Introduction}

\begin{figure}
     \centering
     \begin{subfigure}[b]{\linewidth}
         \centering
         \caption*{\footnotesize \hspace{0.25cm} Input \hspace{0.5cm}  SPatchGAN \hspace{0.2cm}  U-GAT-IT \hspace{0.2cm}  Council-GAN \hspace{0.1cm}  ACL-GAN}
         \includegraphics[width=\linewidth]{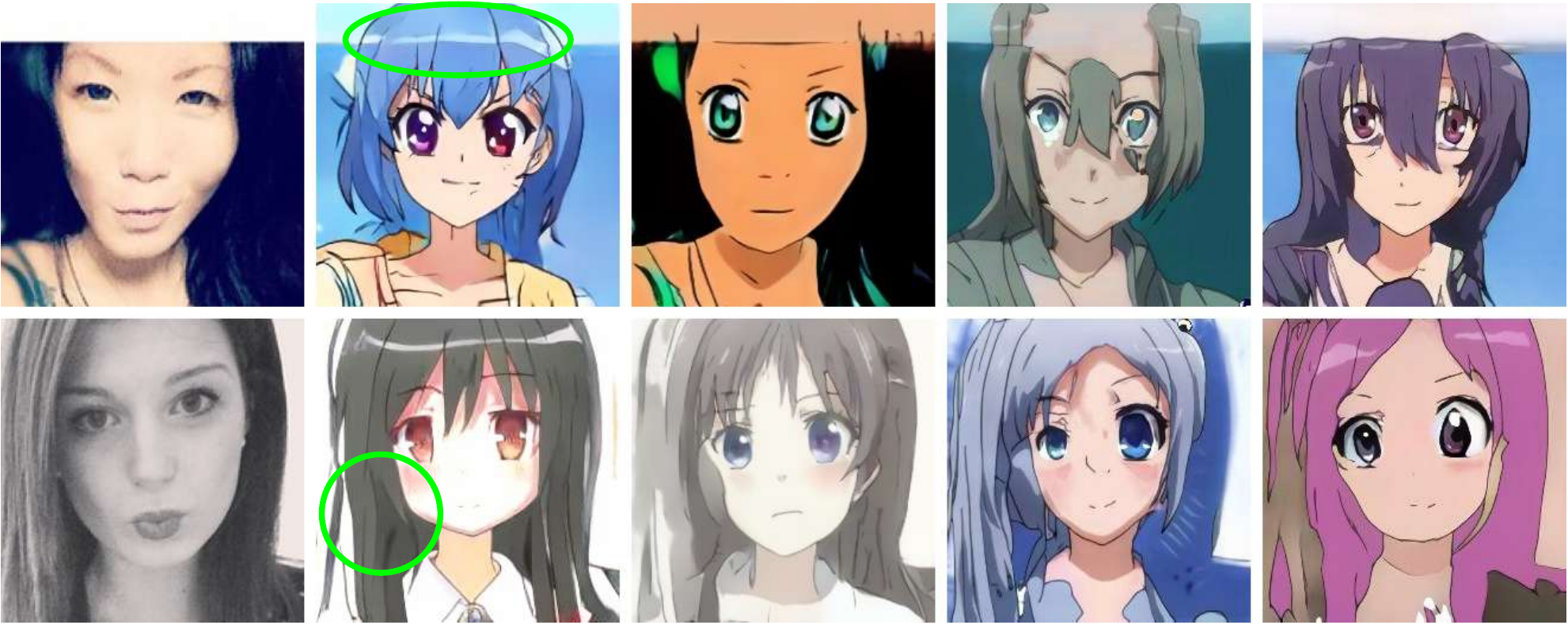}
         \caption{Selfie-to-anime}
         \label{fig:s2a_intro}
    \end{subfigure}
    \hfill
    \begin{subfigure}[b]{\linewidth}
         \centering
         \caption*{\footnotesize \hspace{0.25cm} Input \hspace{0.5cm}  SPatchGAN \hspace{0.2cm}  U-GAT-IT \hspace{0.2cm}  Council-GAN \hspace{0.1cm}  ACL-GAN}
         \includegraphics[width=\linewidth]{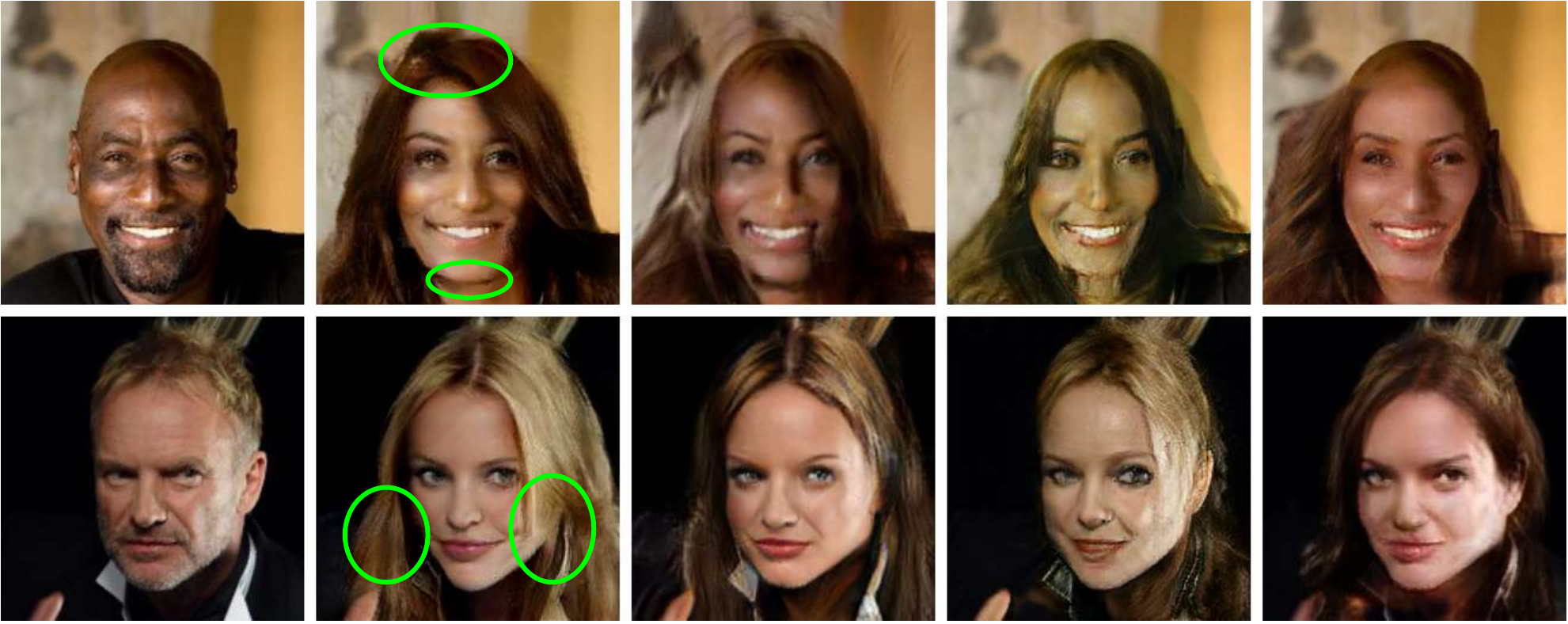}
         \caption{Male-to-female}
         \label{fig:m2f_intro}
     \end{subfigure}
     \hfill
     \begin{subfigure}[b]{\linewidth}
         \centering
         \caption*{\footnotesize \hspace{0.25cm} Input \hspace{0.5cm}  SPatchGAN \hspace{0.2cm}  U-GAT-IT \hspace{0.2cm}  Council-GAN \hspace{0.1cm}  ACL-GAN}
         \includegraphics[width=\linewidth]{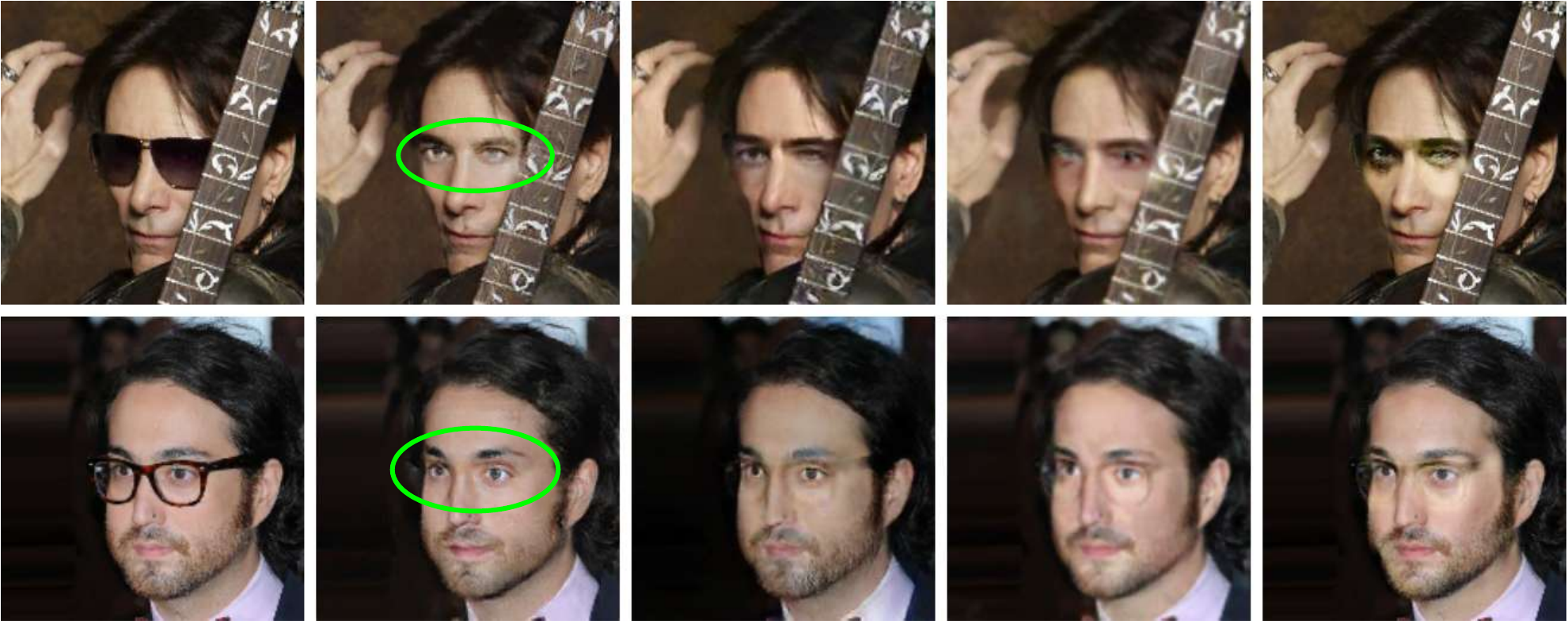}
         \caption{Glasses removal}
         \label{fig:glasses_intro}
     \end{subfigure}
\caption{Example results for our SPatchGAN and the baselines. Our method generates natural looking hairstyle in Figure \ref{fig:s2a_intro} and Figure \ref{fig:m2f_intro}, and suppresses the traces of glasses in Figure \ref{fig:glasses_intro}. The improvements in shape deformation are highlighted with green circles.} 
\label{fig:cmp_with_sota_intro}
\end{figure}

Unsupervised image-to-image translation has become an area of growing interest in computer vision. Powered by generative adversarial networks (GANs) \cite{DBLP:conf/nips/GoodfellowPMXWOCB14}, recent works \cite{DBLP:conf/iccv/ZhuPIE17, DBLP:conf/icml/KimCKLK17, DBLP:conf/nips/LiuBK17, DBLP:conf/iclr/TaigmanPW17, DBLP:conf/icml/AlmahairiRSBC18, DBLP:conf/eccv/HuangLBK18, DBLP:conf/cvpr/ChenLL18, DBLP:conf/iccv/SiddiqueeZTFGBL19, DBLP:journals/ijcv/LeeTMHLSY20} are able to change the local texture and style of the images without the assistance of any paired data. However, these methods still have difficulty with tasks which require a larger shape deformation. Our goal is to improve unsupervised image-to-image translation for applications which involve geometric changes between the source and the target domains.

GANs are trained by a game between a generator network and a discriminator network. At the core of the modern unsupervised image translation frameworks \cite{DBLP:conf/iccv/ZhuPIE17, DBLP:conf/eccv/HuangLBK18, DBLP:conf/iclr/KimKKL20, DBLP:conf/eccv/ParkEZZ20, DBLP:conf/cvpr/NizanT20, DBLP:conf/eccv/ZhaoWD20} is a PatchGAN discriminator \cite{DBLP:conf/cvpr/IsolaZZE17}. Instead of giving a single real vs. fake probability for the whole image, PatchGAN produces multiple outputs for multiple overlapping image patches. It helps with the convergence of the network by involving less parameters and stopping at a relatively low-level scale \cite{DBLP:conf/cvpr/IsolaZZE17}.

PatchGAN is feasible for identifying a specific type of texture all over the image, since it processes all the patches by the same set of convolutional weights. However, we find it less capable of dealing with complicated shape changes like selfie-to-anime. In such a case, the discriminative features for one area, \eg., the eyes of an anime face, can be very different from another area, \eg., the hair. Therefore, it becomes difficult to make an accurate and stable decision for every patch.

A related problem is that PatchGAN alone cannot guarantee the stability of an unsupervised image translation network. Various constraints have been proposed to stabilize and improve the training process, including the cycle constraints \cite {DBLP:conf/iccv/ZhuPIE17, DBLP:conf/icml/KimCKLK17} and the shared latent space assumptions \cite {DBLP:conf/nips/LiuBK17, DBLP:conf/eccv/HuangLBK18}. The constraints are applied on the generator to help alleviate some typical problems such as mode collapse. However, these methods often create a conflict between the interest of GANs and the additional constraints, resulting in incompletely translation. \Eg, it is challenging to reshape the hair properly in Figure \ref{fig:s2a_intro} and Figure \ref{fig:m2f_intro}, and remove the traces of glasses in Figure \ref{fig:glasses_intro}.

The above issues can be potentially solved if one can stabilize the discriminator itself instead of introducing more constraints on the generator. With this motivation, we propose SPatchGAN, an improved multi-scale discriminator architecture which leverages the statistical features over patches. The network benefits from the stability and global view of statistical features at the low-level scales, and the strong discrimination capability at the high-level scales. All the scale levels are jointly optimized in a unified network. With the improved discriminator, we are able to reduce the full cycle constraints to a weak form that only operates on the forward cycle and the low resolution images. 

The experiments demonstrate that our method is superior to the existing state-of-the-art methods. The main contributions are:
\begin{itemize}
\setlength\itemsep{-0.2em}
  \item Our novel discriminator architecture stabilizes the network by matching the distributions of the statistical features, and employing a shared backbone for the multiple scales.
  \item We propose a simplified framework that reduces the conflicts between GANs and the other constraints, and facilitates the shape deformation.
\end{itemize}

%% file: sections/related_work.tex
\section{Related Work}

\textbf{GANs.} GANs have provided a powerful tool for matching two distributions, and delivered promising results for various applications. However, GANs are usually non-ideal in practice due to some stability issues such as mode collapse and oscillation \cite{DBLP:conf/iclr/MetzPPS17}. 

Various methods \cite{DBLP:conf/nips/SalimansGZCRCC16, DBLP:conf/iclr/MetzPPS17, DBLP:journals/corr/ArjovskyCB17, DBLP:conf/nips/GulrajaniAADC17, DBLP:conf/iccv/MaoLXLWS17, DBLP:conf/iclr/MiyatoKKY18} have been studied to stabilize GANs. Our work is inspired by the feature matching technique \cite{DBLP:conf/nips/SalimansGZCRCC16} which aligns the features of the generated images with the real images at an intermediate layer. We enhance the feature matching technique by considering distribution matching instead of minimizing the L1 or L2 distance. Moreover, our method operates on the statistics of activations instead of individual activation values. We also extend the method to multiple scales.

Many studies \cite{DBLP:journals/corr/MirzaO14, DBLP:journals/corr/RadfordMC15, DBLP:conf/iclr/KarrasALL18, DBLP:conf/icml/ZhangGMO19, DBLP:conf/iclr/BrockDS19, DBLP:conf/cvpr/KarrasLA19, DBLP:conf/cvpr/KarnewarW20, DBLP:conf/cvpr/KarrasLAHLA20} are related to improving GANs for random image generation, but these methods are not always applicable to image translation. \Eg, MSG-GAN \cite{DBLP:conf/cvpr/KarnewarW20} facilitates the flow of gradients by connecting the matching layers of the generator and discriminator, but there is no such layer correspondence in the typical image translation networks. Therefore, it is necessary to specifically optimize the network architecture for image translation.

\textbf{Discriminator Structures for Image Translation.} The discriminator in the original GAN framework \cite{DBLP:conf/nips/GoodfellowPMXWOCB14} is simply a binary classifier. For \emph{supervised} image translation, Isola \etal \cite{DBLP:conf/cvpr/IsolaZZE17} propose the PatchGAN discriminator to classify if each image patch is real or fake. The responses of all the patches are averaged to provide the final output. PatchGAN is initially designed for improving the high-frequency part of the generated image, while the correctness of the overall structure is guaranteed by the supervision signal.

The PatchGAN structure has been extended to multiple scales \cite{DBLP:conf/cvpr/Wang0ZTKC18} to cover the low frequency part as well, and has been widely adopted by the latest \emph{unsupervised} image translation networks \cite{DBLP:conf/eccv/HuangLBK18, DBLP:conf/iclr/KimKKL20, DBLP:conf/cvpr/NizanT20, DBLP:conf/eccv/ZhaoWD20}. Among these methods, U-GAT-IT  \cite{DBLP:conf/iclr/KimKKL20} applies PatchGAN on two scales and uses a Class Activation Map (CAM) \cite{DBLP:conf/cvpr/ZhouKLOT16} based attention module. MUNIT \cite{DBLP:conf/eccv/HuangLBK18} and ACL-GAN \cite{DBLP:conf/eccv/ZhaoWD20} collect the outputs from three scales. The multi-scale PatchGAN usually has an independent network for each scale, \eg, the discriminator in \cite{DBLP:conf/eccv/HuangLBK18, DBLP:conf/eccv/ZhaoWD20} consists of three networks respectively for the raw image, the $1/4$ sized image and the $1/16$ sized image. Different from the above methods, we switch our attention from the individual patches to the statistical features over patches, and employ a unified network for multiple scales.

\textbf{Unsupervised Image Translation Frameworks.} A cycle based framework \cite{DBLP:conf/iccv/ZhuPIE17, DBLP:conf/icml/KimCKLK17} is widely used to stabilize GANs for unsupervised image translation. In this framework, two generators are jointly optimized to enforce a forward cycle constraint and a backward cycle constraint. Though the framework is effective in preventing mode collapse, it often causes irrelevant traces of the source image to be left on the generated image \cite{DBLP:conf/cvpr/NizanT20, DBLP:conf/eccv/ZhaoWD20}. The cycle constraints are often used together with an identity mapping constraint \cite{DBLP:conf/iclr/TaigmanPW17} which reconstructs a target domain image with the generator. CUT \cite{DBLP:conf/eccv/ParkEZZ20} achieves a similar goal with cycle constraints through a different approach, enforcing the corresponding elements of the source and generated images to be mapped to a similar feature vector. In our work, the cycle constraints are relaxed to reduce the side effects.

Some recent works \cite{DBLP:conf/eccv/HuangLBK18, DBLP:conf/icml/AlmahairiRSBC18, DBLP:journals/ijcv/LeeTMHLSY20} further consider the problem of generating multiple output images for a given source image. MUNIT \cite{DBLP:conf/eccv/HuangLBK18} supports multimodality by encoding the source image into a content code and a style code. Council-GAN \cite{DBLP:conf/cvpr/NizanT20} further considers distribution matching among multiple generators, while ACL-GAN \cite{DBLP:conf/eccv/ZhaoWD20} considers distribution matching between the cycle output and the identity mapping output. The multimodal frameworks are usually more complicated than the unimodal frameworks, \eg, MUNIT requires the reconstruction of the image from the content and style codes, and the reconstruction of the codes from the image. In this paper we mainly focus on the unimodal case for simplicity, but we also compare our results to the the state-of-the-art multimodal methods. The applicability of our method in multimodal translation is studied in the supplementary materials.

%% file: sections/model.tex
\section{Model}

Our model consists of a forward generator $G$, an SPatchGAN discriminator $D$, and a backward generator $B$ that operates on the low resolution images. The forward generator translates a source image $x_1$ to a generated image $G(x_1)$. $G$ and $D$ play an adversarial game, in which $D$ aims to distinguish the generated images from the real images, while $G$ aims to fool the discriminator. $B$ is jointly optimized with $G$ to ensure the similarity between the source image and the generated image, and to stabilize the network.

The source domain and the target domain for the image translation task are denoted $\mathcal{X}_1$ and $\mathcal{X}_2$. We denote the distribution of source images $x_1 \in \mathcal{X}_1$ as $p^{src}(x)$, the distribution of real images $x_2 \in \mathcal{X}_2$ as $p^{data}(x)$, and the distribution of the generated images $G(x_1)$ as $p^{g}(x)$.  The goal is to match $p^{g}(x)$ with $p^{data}(x)$, and preserve the common characteristics between the two domains during the translation.

\input{model/dis_structure}
\input{model/analysis_dis}
\input{model/training_framework}
\input{model/implementation}

%% file: model/dis_structure.tex
\subsection{Discriminator Structure}

\begin{figure}[t]
\centering
\includegraphics[width=\linewidth]{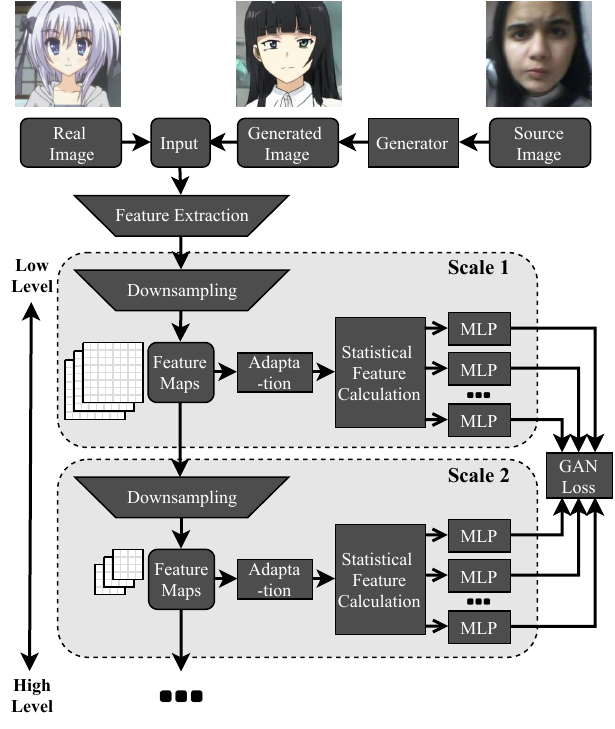}
\caption{Our multi-scale discriminator structure. The feature maps of each scale are processed by an adaptation block, a statistical feature calculation block and $N$ MLPs to generate $N$ outputs.}
\label{fig:statsgan_dis}
\end{figure}

Our discriminator structure is shown in Figure \ref{fig:statsgan_dis}. $D$ takes an image $x$ as the input, which can be either a real image $x_2$ or a generated image $G(x_1)$. $D$ produces multiple outputs $D_{m, n}(x)$, $m=1,2,...M$, $n  = 1,2,...N$, where $M$ is the number of scales, $N$ is the number of statistical features. $D$ has an initial feature extraction block, followed by $M$ scales. Each scale consists of a downsampling block, an adaptation block, a statistical feature calculation block and $N$ multilayer perceptrons (MLPs). We show only two scales in Figure \ref{fig:statsgan_dis} for simplicity, starting from the scale at the lowest level, \ie, scale 1.

The feature maps of each scale are generated by the corresponding downsampling block, and are reused by the higher level scales. They are processed by an adaptation block to adjust the features specifically for the current scale. The adapted feature maps at the $m$-th scale are denoted as $h_m(x) \in \mathbb{R}^{H_m \times W_m \times C_m}$, where $H_m$, $W_m$ and $C_m$ are the height, width and number of channels of the adapted feature maps. $h_m$ is the function for the backbone network of the $m$-th scale. It is jointly defined by the initial feature extraction block, the downsampling blocks up to the $m$-th scale and the adaptation block at the $m$-th scale. Each element of the adapted feature maps corresponds to a patch of the input image. A higher level has a larger patch size.

The adapted feature maps are sent to the statistical feature calculation block to generate the channel-wise statistics. For the $m$-th scale, $N$ statistical feature vectors $s_{m,n}(x) \in \mathbb{R}^{C_m}$, $n  = 1,2,...N$ are calculated,
\begin{equation}
\label{eqn:stats}
s_{m,n}(x) = g_n(h_m(x)).
\end{equation}
$g_n$ is the function that calculates the $n$-th statistical feature.

Each statistical feature vector is further processed by a multilayer perceptron (MLP) to finally generate a scalar output,
\begin{equation}
\label{eqn:mlp}
D_{m,n}(x) = f_{m,n}(s_{m,n}(x)),
\end{equation}
where $f_{m,n}$ is defined by the MLP for the $n$-th statistical feature at the $m$-th scale.

Different from the individual patch based methods, the output of our discriminator is derived from the statistical features $s_{m,n}(x)$. The statistical features are usually more stable than the individual patches, since each of their elements is calculated over $H_mW_m$ patches. In contrast to the limited perceptive field of individual patches, every statistical feature is about the whole image. The global view is helpful for verifying the correctness of shape deformation, \eg, the change of the hairstyle. Furthermore, we employ a dedicated MLP for each statistical feature, rather than the shared convolutional weights in PatchGAN. This makes each output of our discriminator more accurate.

%% file: model/analysis_dis.tex
\subsection{Analysis of Statistical Feature Matching}

Our discriminator is designed to match the statistical features of the generated images with the real images. At a high-level scale, matching the statistical features is similar to fully matching the input distributions, since the high-level features are very discriminative. It also implies the difficulty for achieving a perfect matching at the high level. However, we can expect to match the statistical features at the lower levels with more confidence.

An important assumption in the original GAN framework is training $D$ until optimal after each update of $G$ \cite{DBLP:conf/nips/GoodfellowPMXWOCB14}. For a given input $x$, the optimal output is 
\begin{equation}
\label{eqn:optimal_D}
D^*(x)=\frac{p^{data}(x)}{p^{data}(x)+p^g(x)}.
\end{equation}
However, this assumption is usually not met in practice, where $D$ is typically a complicated neural network to deal with a high dimensional input $x$. It is very challenging to train $D$ until optimal due to issues such as the saddle points and the local minimum.

Instead of directly matching the distributions of the input images, our target is matching the distributions of the statistical features $s_{m,n}(x)$ in Eq. \ref{eqn:stats}. We consider $s_{m,n}(x)$ as a \emph{feature input}, and its corresponding MLP $f_{m,n}$ as a \emph{feature discriminator}.
$s_{m,n}(x)$ as an input has a much lower dimension than the original input $x$, so a simple MLP can serve as its discriminator with enough capacity. This makes it easier to train the feature discriminator $f_{m,n}$ to optimal. Similar to Eq. \ref{eqn:optimal_D}, the optimal output of the feature discriminator is
\begin{equation}
\begin{aligned}
\label{eqn:stats_ratio_est}
D_{m,n}(x) &= f^*_{m,n}(s_{m,n}(x)) \\ 
&= \frac{p^{data}_{m,n}(s_{m,n}(x))}{p^{data}_{m,n}(s_{m,n}(x))+p^g_{m,n}(s_{m,n}(x))},
\end{aligned}
\end{equation}
where $p^{data}_{m,n}(s)$ and $p^g_{m,n}(s)$ are the distributions of $s_{m,n}(x)$ for the real and generated images respectively. We expect $D_{m,n}(x)$ to be close to $0.5$ when the two distributions are well matched.

On the other hand, there is an disadvantage of the feature input $s_{m,n}(x)$. Unlike the original input image, the feature input additionally depends on a part of the discriminator $h_m$ according to Eq. \ref{eqn:stats}. As a result, the feature discriminator is simultaneously updated with its input distributions $p^{data}_{m,n}(s)$ and $p^g_{m,n}(s)$, even if the generator is fixed. The main challenge for the feature discriminator turns out to be that its knowledge about the feature inputs is not fully up to date. Put simply, the feature discriminator is non-ideal unless $h_m$ is fixed.

As the backbone of the discriminator, $h_m$ is optimized to make the statistical features discriminative. We cannot fix $h_m$, but we can stabilize $h_m$ as much as possible. This is where our network structure plays an important role. The layers in $h_m$ are shared by multiple scales and multiple outputs, so they are unlikely to be changed drastically according to one output. $h_m$ at a low-level scale is especially stable, since it contains less layers, and receives optimization signals from more higher level outputs. Moreover, the statistical features are further stabilized by the calculation over multiple patches, especially at a low-level scale where the number of patches is large.

As long as the feature inputs are slow varying compared to the feature discriminator, the output of the discriminator becomes relatively accurate. As a result, $G$ can be optimized in the correct direction to match $p^g_{m,n}(s)$ with $p^{data}_{m,n}(s)$ according to the signals from $D$. Though the distributions are dependent on the non-cooperative $h_m$, chances are they can be matched by $G$ when $h_m$ is stable enough.

Therefore, a good match between low-level statistical features can be practically achieved using the discriminator designed in our fashion, encouraging the generator to avoid the potential flaws that would break the distribution matching. Our experiments demonstrate that such statistical feature matching mechanism effectively alleviates mode collapse and stabilizes the training process. 

%% file: model/training_framework.tex
\subsection{Training Framework}

Our training framework has three objectives. An adversarial loss is employed for matching the distributions of the statistical features. A weak cycle loss and an identity loss are added to keep the generated image correlated with the input image, and to further stabilize the network.

\textbf{Adversarial Loss.} We adopt the Least Squares GANs (LSGANs) objective \cite{DBLP:conf/iccv/MaoLXLWS17} instead of the original GAN objective \cite{DBLP:conf/nips/GoodfellowPMXWOCB14} for stable training. We use the binary 0-1 coding variant to keep the optimal output of the discriminator unchanged. The loss function for $D$ is averaged over multiple scales and multiple statistical features,
\begin{equation}
\begin{split}
\label{eqn:adv_loss_dis}
L^{adv}_D = \frac{1}{MN}\sum_m\sum_n( \mathbb{E}_{x \sim p^{data}(x)}[(D_{m,n}(x)-1)^2] \\ +  \mathbb{E}_{x \sim p^{src}(x)}[D_{m,n}(G(x))^2]).
\end{split}
\end{equation}
Similarly, the adversarial loss function for $G$ is
\begin{equation}
\begin{split}
\label{eqn:adv_loss_gen}
L^{adv}_G = \frac{1}{MN}\sum_m\sum_n( \mathbb{E}_{x \sim p^{src}(x)}[(D_{m,n}(G(x))-1)^2]).
\end{split}
\end{equation}

\textbf{Weak Cycle Loss.} In the original cycle based framework, a forward cycle constraint $x_1=B(G(x_1))$ and a backward cycle constraint $x_2=G(B(x_2))$ are necessary for preventing mode collapse \cite{DBLP:conf/iccv/ZhuPIE17}. After improving the discriminator, we find the network stable enough with the forward cycle only. We simply remove the backward cycle, and thus reduce the complexity by half.

Moreover, we further simplify the backward generator $B$, \ie, the generator from the target domain to the source domain. Compared to the original backward generator which operates on the full sized images, ours reconstructs a low resolution source image from a low resolution generated image. The weak cycle loss function is
\begin{equation}
\label{eqn:weak_cycle_loss}
L^{cyc} = \mathbb{E}_{x \sim p^{src}(x)}[\lVert u(x) - B(u(G(x))) \rVert_1],
\end{equation}
where $u(x)$ is the operation to resize a full sized image to a low resolution image.

Since the weak cycle constraint is applied on the low resolution images, only the low frequency information of the source image is required to be preserved in the generated image. Compared to the full cycle constraints, the weak cycle constraint suppresses the leftovers from the source image, and facilitates the shape deformation. On the other hand, it still helps to keep the overall structure of the source image, such as the pose of face in the selfie-to-anime case. We know that some high frequency information may also need to be preserved depending on the application, and leave this job to the identity loss. 

\textbf{Identity Loss.} Our identity loss is similar to that in \cite{DBLP:conf/iccv/ZhuPIE17, DBLP:conf/iclr/KimKKL20}, with the exception that the loss is applied to the target domain only. Unlike the weak cycle loss, the identity loss is applied on the full sized images. The identity loss function is
\begin{equation}
\label{eqn:identity_loss}
L^{id} = \mathbb{E}_{x \sim p^{data}(x)}[\lVert x - G(x) \rVert_1].
\end{equation}
The identity loss turns the generator into an autoencoder for the target domain. This helps to preserve the common parts in the source and the target domains, such as the background in the glasses removal case. Compared to the cycle loss, the identity loss is less likely to conflict with the adversarial loss, since the input for the generator is taken from a different domain.

\textbf{Full Objective.} The total loss of the forward and backward generators is the weighted sum of Eq. \ref{eqn:adv_loss_gen}, Eq. \ref{eqn:weak_cycle_loss} and Eq. \ref{eqn:identity_loss}, 
\begin{equation}
\label{eqn:gen_objective}
L_G = \lambda^{adv}L^{adv}_G + \lambda^{cyc}L^{cyc} + \lambda^{id}L^{id},
\end{equation}
The objective of the generators is minimizing $L_G$. The objective of the discriminator is minimizing its loss in Eq. \ref{eqn:adv_loss_dis}.

%% file: model/implementation.tex
\subsection{Implementation}

We describe the basic structure and parameter settings of our model in this section. The details are available in the supplementary materials. The height and width of the input images are assumed to be $H_0$ and $W_0$.

\textbf{Discriminator.} We set the number of scales to be $M=4$. The height and width of the feature maps are downsampled to $\frac{H_0}{4}$ and $\frac{W_0}{4}$ by the feature extraction block, and further downsampled by a factor of 2 in each scale. The convolutional layers for downsampling are similar to those in the multi-scale PatchGAN \cite{DBLP:conf/eccv/HuangLBK18, DBLP:conf/cvpr/NizanT20, DBLP:conf/eccv/ZhaoWD20}. The adaptation block consists of two $1 \times 1$ convolutional layers. We use spectral normalization \cite{DBLP:conf/iclr/MiyatoKKY18} in the discriminator.

Our discriminator is flexible enough to incorporate with various kinds of  statistical features. In our experiments, we find that a good performance can be achieved with three basic ones, \ie, $N=3$. The function $g_1$ in Eq. \ref{eqn:stats} calculates the channel-wise mean values using global average pooling. $g_2$ calculates the channel-wise maximum values using global max pooling. $g_3$ calculates the channel-wise uncorrected standard deviation for each feature map. In our implementation, the statistical feature calculation layers are pre-defined rather than learnable, \ie, the contributions of all patches are taken into account in a fixed manner. In contrast, a learnable channel-wise
reduction layer may dynamically switch its attention on different patches during the training process, making the calculated features less robust.

\textbf{Generator.} The generator in our framework can be chosen independently of the discriminator. In our implementation, we adopt a modified version of CycleGAN \cite{DBLP:conf/iccv/ZhuPIE17} as our forward and backward generator. To further facilitate shape deformation, we apply the residual blocks to feature maps of size $\frac{H_0}{8} \times \frac{W_0}{8}$ instead of $\frac{H_0}{4} \times \frac{W_0}{4}$. Accordingly, the input dimensions of the backward generator are resized to $\frac{H_0}{8} \times \frac{W_0}{8}$. We also use layer normalization \cite{DBLP:journals/corr/BaKH16} instead of instance normalization \cite{DBLP:conf/cvpr/UlyanovVL17} in the upsampling layers to alleviate the blob artifact problem \cite{DBLP:conf/cvpr/KarrasLAHLA20}.

%% file: sections/experiments.tex
\section{Experiments}

\input{experiments/experiment_setup}
\input{experiments/ablation}
\input{experiments/cmp_baseline}

%% file: experiments/experiment_setup.tex
\subsection{Experiment Setup}

\textbf{Datasets.} Similar to \cite{DBLP:conf/cvpr/NizanT20, DBLP:conf/eccv/ZhaoWD20}, we evaluate SPatchGAN on three tasks including selfie-to-anime, male-to-female and glasses removal.

\emph{Selfie-to-Anime.} The selfie-to-anime dataset \cite{DBLP:conf/iclr/KimKKL20} contains 3,400 / 100 selfie images and 3,400 / 100 anime face images in the training / test set. The image size is $256\times256$.

\emph{Male-to-Female.} The male-to-female dataset \cite{DBLP:conf/cvpr/NizanT20} contains face images cropped from CelebA \cite{DBLP:conf/iccv/LiuLWT15}. The training / test set contains 68,261 / 16,173 images of male, and 94,509 / 23,656 images of female. The image size is $218\times178$.

\emph{Glasses Removal.} The original glasses removal dataset \cite{DBLP:conf/cvpr/NizanT20} contains face images cropped from CelebA \cite{DBLP:conf/iccv/LiuLWT15} for both male and female. It has a data imbalance problem that there are much more images of male with glasses than female with glasses. As a side effect, the face may become more feminine after removing the glasses \cite{DBLP:conf/cvpr/NizanT20, DBLP:conf/eccv/ZhaoWD20}. To avoid this problem and focus on the glasses removal task, we only use the male images in our experiments. The training / test set contains 8,366 / 2,112 images of male with glasses, and 59,895 / 14,061 images of male without glasses. The image size is the same as male-to-female.

\textbf{Baseline Models.} We compare our SPatchGAN to the state-of-the-art models for unsupervised image translation, including CycleGAN \cite{DBLP:conf/iccv/ZhuPIE17}, MUNIT \cite{DBLP:conf/eccv/HuangLBK18}, U-GAT-IT \cite{DBLP:conf/iclr/KimKKL20}, CUT \cite{DBLP:conf/eccv/ParkEZZ20}, Council-GAN \cite{DBLP:conf/cvpr/NizanT20} and ACL-GAN \cite{DBLP:conf/eccv/ZhaoWD20}. CycleGAN and CUT uses a single scale PatchGAN discriminator, while the others use the multi-scale PatchGAN. CycleGAN and U-GAT-IT are based on the cycle constraints. CUT is based on a contrastive learning approach which maximizes the mutual information between the source and generated images. MUNIT, Council-GAN and ACL-GAN are based on a partially shared latent space assumption.  We use the official pre-trained models if available, including the selfie-to-anime and male-to-female models of Council-GAN, and the selfie-to-anime model of U-GAT-IT. The other results are reproduced using the official code base.

\textbf{Evaluation Metrics.} We adopt two metrics, the Fr\'echet Inception Distance (FID) \cite{DBLP:conf/nips/HeuselRUNH17} and the kernel inception distance (KID) \cite{DBLP:conf/iclr/BinkowskiSAG18}, for the quantitative evaluation. FID is a widely used metric for comparing the distributions of the real and generated images. KID is an improved metric which takes additional aspects of the distributions into account, and is unbiased.

\textbf{Training.} Our model is trained with $\lambda^{adv}=4$ and $\lambda^{id}=10$ for all the datasets. $\lambda^{cyc}$ is set to $20$, $10$ and $30$ respectively for selfie-to-anime, male-to-female and glasses removal.

We use an Adam \cite{DBLP:journals/corr/KingmaB14} optimizer with $\beta_1=0.5$ and $\beta_2=0.999$. We also use a weight decay at rate of 0.0001. The models are trained for 500k iterations with a batch size of four. The learning rate is 0.0001 for the first 100k iterations and linearly decayed to 0.00001. The training takes about 2 hours per 10k iterations on a NVIDIA Tesla V100 GPU. As a reference, U-GAT-IT \cite{DBLP:conf/iclr/KimKKL20} in its light mode costs about 3.5 hours per 10k iterations with the same batch size and infrastructure. The speed-up of our method is from the simplified training framework.

%% file: experiments/ablation.tex
\subsection{Ablation Studies}

We compare SPatchGAN to several variants, including 1) removing the channel-wise mean feature, 2) removing the channel-wise maximum feature. 3) removing the channel-wise standard deviation feature, 4) replacing the whole SPatchGAN discriminator with the multi-scale PatchGAN discriminator in \cite{DBLP:conf/eccv/HuangLBK18, DBLP:conf/eccv/ZhaoWD20}.

\begin{figure}
\centering
\caption*{\footnotesize \hspace{0.2cm} Input \hspace{0.25cm} SPatchGAN \hspace{0.1cm} w/o Mean \hspace{0.1cm}  w/o Max \hspace{0.1cm}  w/o Stddev \hspace{0.1cm}  PatchGAN}
\includegraphics[width=\linewidth]{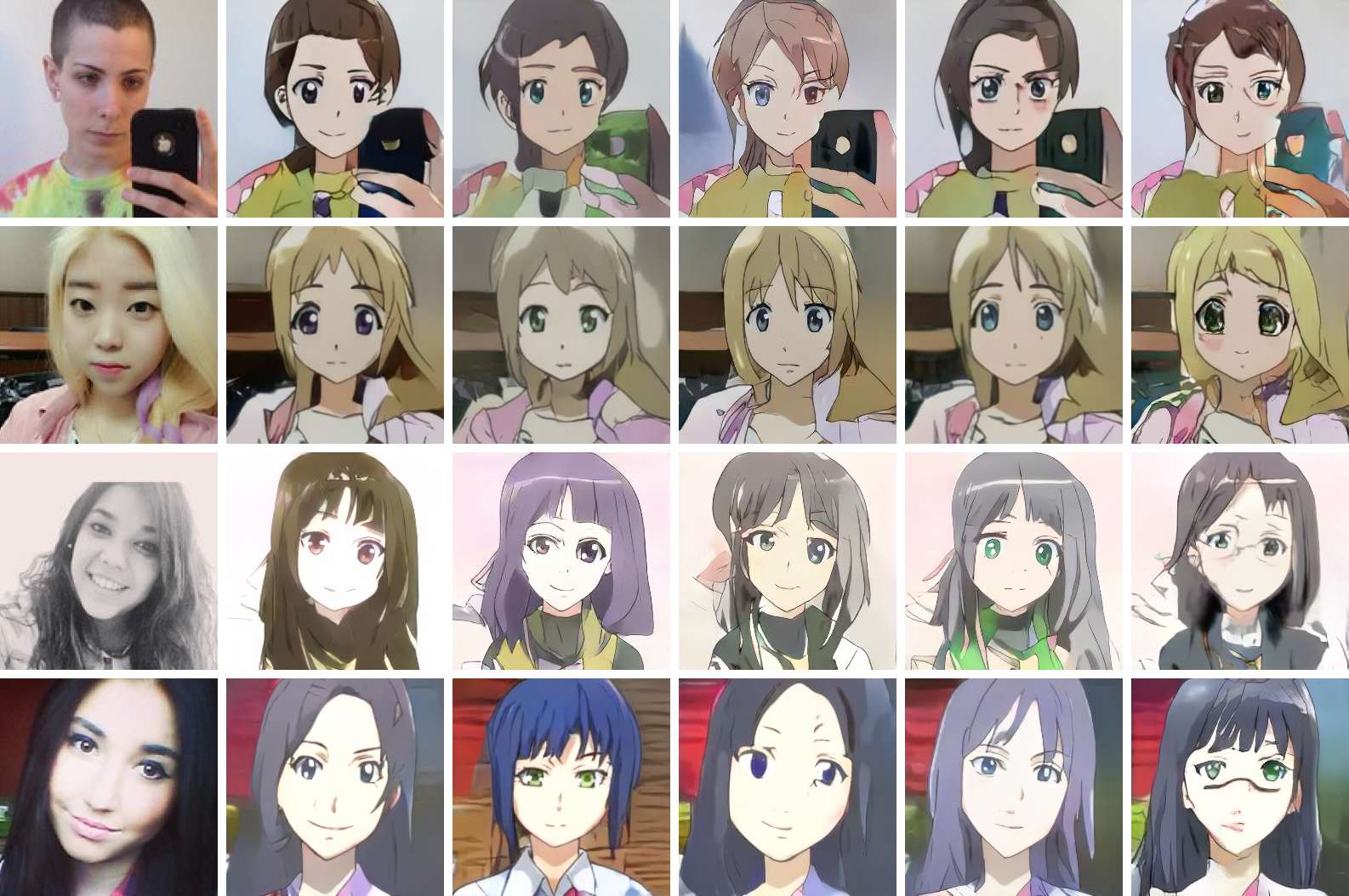}
\caption{Generated images of the ablation study cases for selfie-to-anime.}
\label{fig:s2a_ablation}
\end{figure}

The generated images are shown in Figure \ref{fig:s2a_ablation}. We observe some minor defects when one of the statistical features is removed, \eg, inconsistent color of the two eyes, redundant lines on the face, and blurred local texture. The multi-scale PatchGAN often generates collapsed eyes, indicating that it failed to ensure the correctness of the individual patches when the constraints on the generator are relaxed.

The quantitative results are summarized in Table \ref{tab:ablation}. As a reference, we also include the average FID and KID of the real images to demonstrate the performance in the ideal case. For the real image case, the training images of the target domain are compared to the testing images of the target domain. For the other cases, the generated images are compared to the testing images of the target domain. SPatchGAN has the best FID and KID scores, since it considers the full set of statistical features that reflect the key aspects of the distribution. However, there is still a gap between SPatchGAN and the real images, implying the difficulty of full distribution matching. The performance slightly deteriorate when one of the statistical features is removed. The multi-scale PatchGAN performs much worse than the SPatchGAN based methods in terms of both FID and KID.

\begin{table}
\begin{center}
\begin{tabular}{|c|c|c|}
\hline
Model                & FID & KID \\ \hline \hline
SPatchGAN            & \textbf{83.3}   & \textbf{0.0214}   \\ \hline
SPatchGAN w/o Mean   & 84.1   & 0.0228   \\ \hline
SPatchGAN w/o Max    & 83.8   & 0.0223   \\ \hline
SPatchGAN w/o Stddev & 84.9   & 0.0214   \\ \hline
Multi-scale PatchGAN             & 94.0   & 0.0362   \\ \hline
\emph{Real images}          & \emph{76.7}   & \emph{0.0030}   \\ \hline
\end{tabular}
\caption{Quantitative results of the ablation study cases for selfie-to-anime. Lower is better.}
\label{tab:ablation}
\end{center}
\end{table}

\begin{figure*}
\centering
\includegraphics[width=1.0\linewidth]{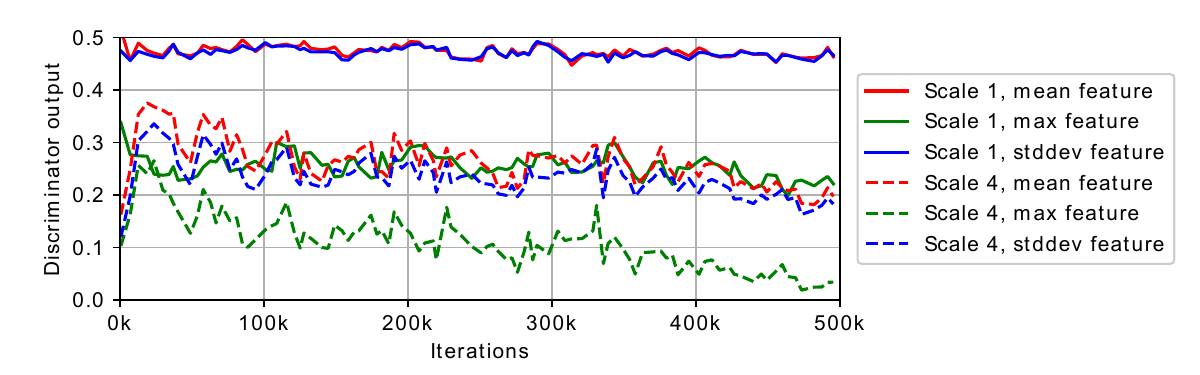}
\caption{The discriminator outputs of SPatchGAN for the generated images in the selfie-to-anime case.}
\label{fig:D_output}
\end{figure*}

\begin{table*}
\centering
  \begin{tabular}{|c|c|c|c|c|c|c|}
    \hline
    \multirow{2}{*}{Model} &
      \multicolumn{2}{c|}{Selfie-to-Anime} &
      \multicolumn{2}{c|}{Male-to-Female} &
      \multicolumn{2}{c|}{Glasses Removal} \\ \cline{2-7}
    & FID & KID & FID & KID & FID & KID \\
    \hline
    SPatchGAN & \textbf{83.3} & \textbf{0.0214} & \textbf{8.73} & \textbf{0.0056} & \textbf{13.9} & \textbf{0.0031} \\
    \hline
    CycleGAN & 92.4 & 0.0299 & 24.5 & 0.0240 & 19.5 & 0.0110 \\
    \hline
    MUNIT & 96.7 & 0.0331 & 20.8 & 0.0161 & 25.1 & 0.0123 \\
    \hline
    U-GAT-IT & 94.8 & 0.0271 & 21.7 & 0.0201 & 18.1 & 0.0080 \\
    \hline
    CUT & 87.2 & 0.0255 & 15.4 & 0.0132 & 15.5 & 0.0047 \\
    \hline
    Council-GAN & 92.4 & 0.0265 & 14.1 & 0.0120 & 25.9 & 0.0197 \\
    \hline
    ACL-GAN & 98.0 & 0.0285 & 13.5 & 0.0112 & 15.3 & 0.0039 \\
    \hline
    \emph{Real images} & \emph{76.7} & \emph{0.0030} & \emph{1.74} & \emph{0.0} & \emph{9.40} & \emph{0.0001} \\
    \hline
  \end{tabular}
\caption{Quantitative results of SPatchGAN and the baselines. Lower is better.}
\label{tab:vs_baselines}
\end{table*}

\begin{figure*}
\centering
\caption*{\footnotesize \hspace{0.4cm} Input \hspace{0.6cm}  SPatchGAN \hspace{0.4cm}  CycleGAN \hspace{0.5cm} MUNIT \hspace{0.6cm} U-GAT-IT \hspace{0.7cm} CUT \hspace{0.5cm} Council-GAN \hspace{0.1cm} ACL-GAN \hspace{0.1cm} }
\includegraphics[width=0.8\linewidth]{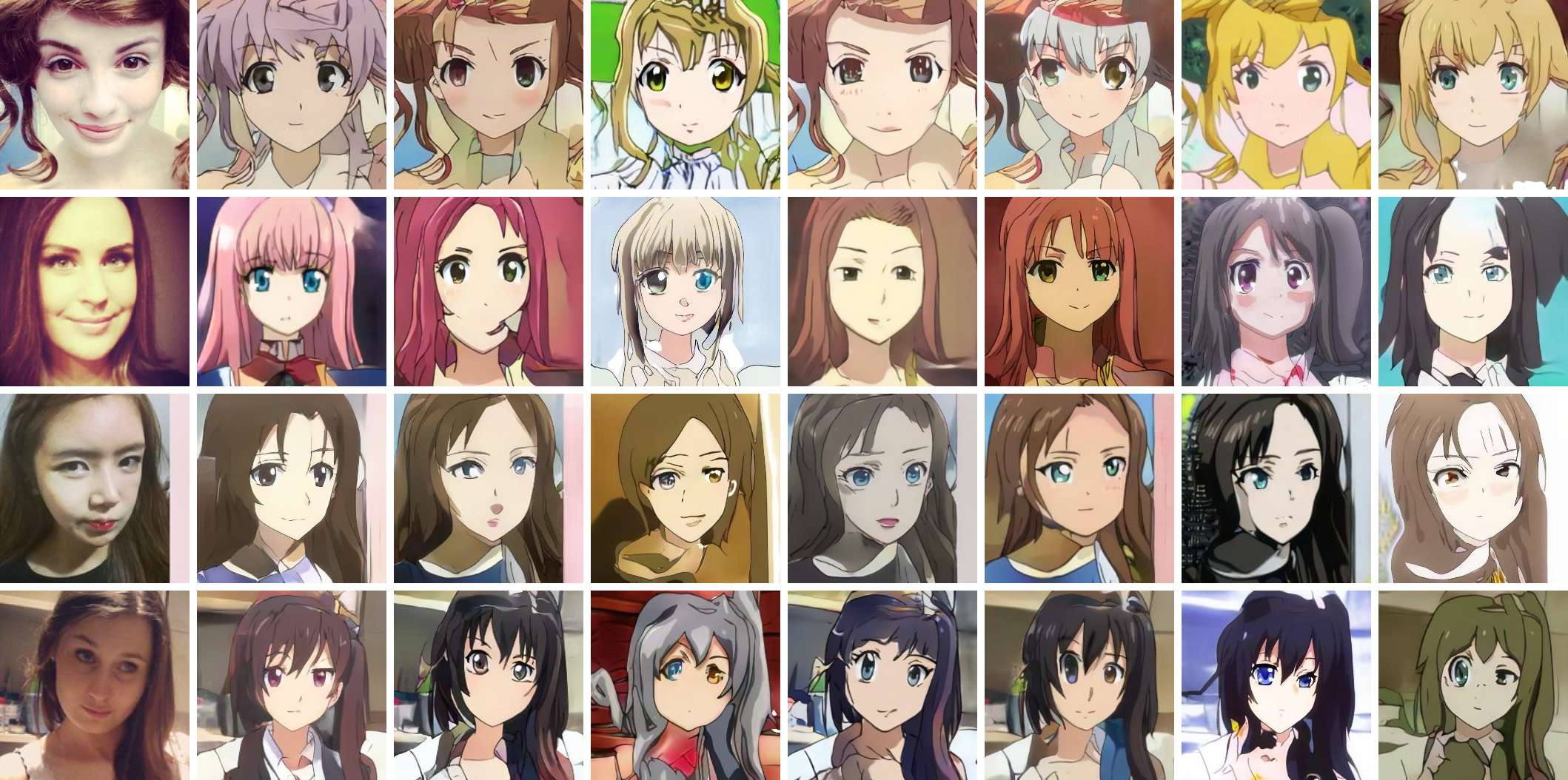}
\caption{Generated images of SPatchGAN and the baselines for selfie-to-anime.}
\label{fig:cmp_s2a}
\end{figure*}

We also take a closer look at the individual outputs of SPatchGAN.
The outputs of the lowest level (scale 1) and the highest level (scale 4) are shown in Figure \ref{fig:D_output} for the generated images. The outputs of the intermediate levels are in between these two. A low output value means that the discriminator can successfully distinguish the generated images from the real images, while an output near 0.5 means that the generated and real images are indistinguishable.

For a given statistical feature, the output value is larger at a lower level. This is aligned with our expectation that it is easier to match the statistical features at the low levels. The outputs at the highest level are much smaller than 0.5, and are consistent with the FID and KID results that there is still some difference between the generated and the real images. The outputs for the channel-wise max feature are below the other two features. This is due to its nature of focusing on the most discriminative patch for each feature channel. It is also worth noting that the channel-wise standard deviation feature is slightly more discriminative than the channel-wise mean feature at the highest level. It implies the importance of considering the relation among patches. In contrast, it is much more difficult for the individual patch based method to be aware of the inter-patch relations.

%% file: experiments/cmp_baseline.tex
\subsection{Comparison with Baseline}

\begin{figure}
\centering
\caption*{\footnotesize \hspace{0.25cm} Input \hspace{0.5cm}  SPatchGAN \hspace{0.2cm}  U-GAT-IT \hspace{0.2cm} Council-GAN \hspace{0.1cm}  ACL-GAN}
\includegraphics[width=\linewidth]{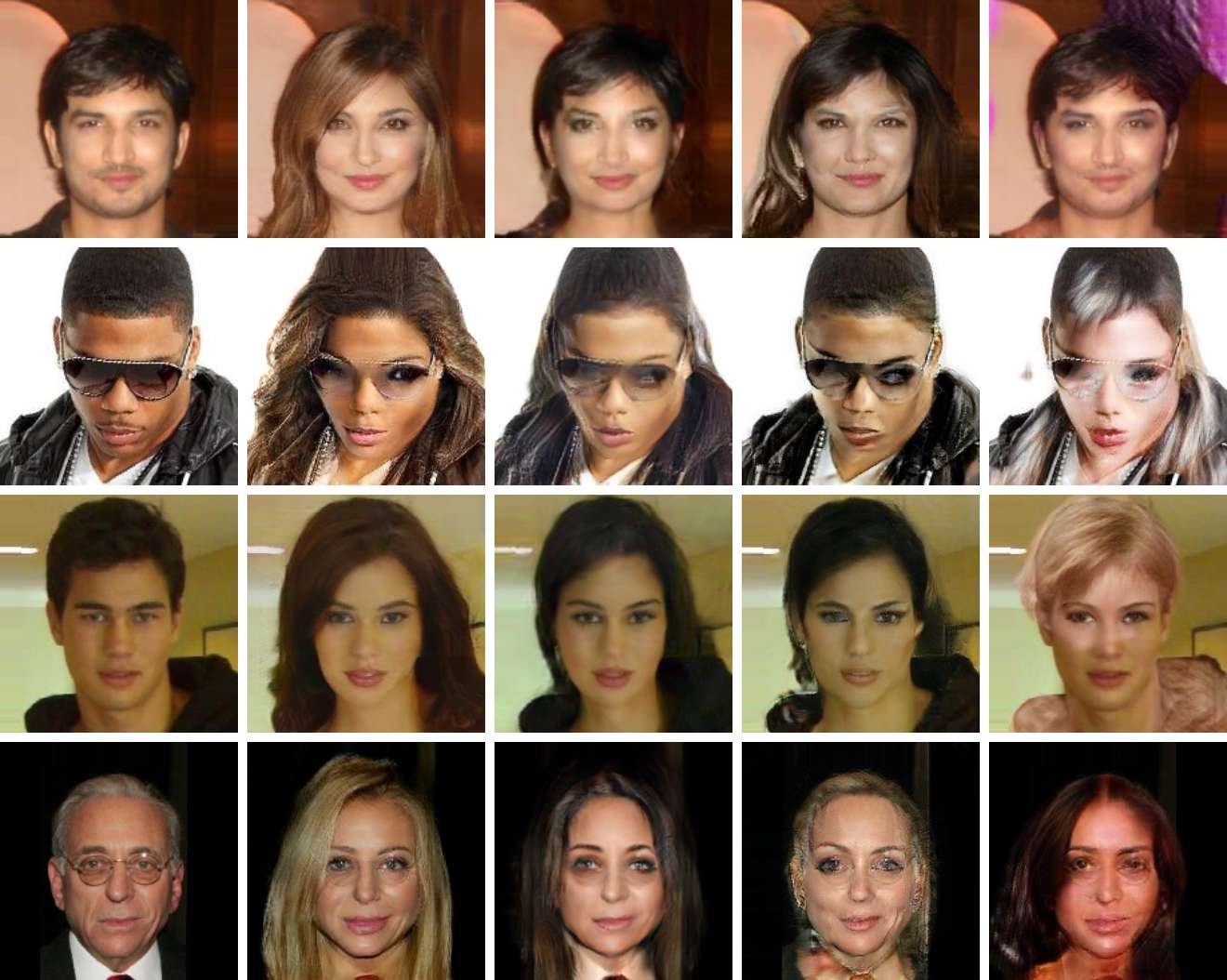}
\caption{Generated images of SPatchGAN and the baselines for male-to-female.}
\label{fig:cmp_m2f}
\end{figure}

\begin{figure}
\centering
\caption*{\footnotesize \hspace{0.25cm} Input \hspace{0.5cm}  SPatchGAN \hspace{0.2cm}  U-GAT-IT \hspace{0.2cm}  Council-GAN \hspace{0.1cm}  ACL-GAN}
\includegraphics[width=\linewidth]{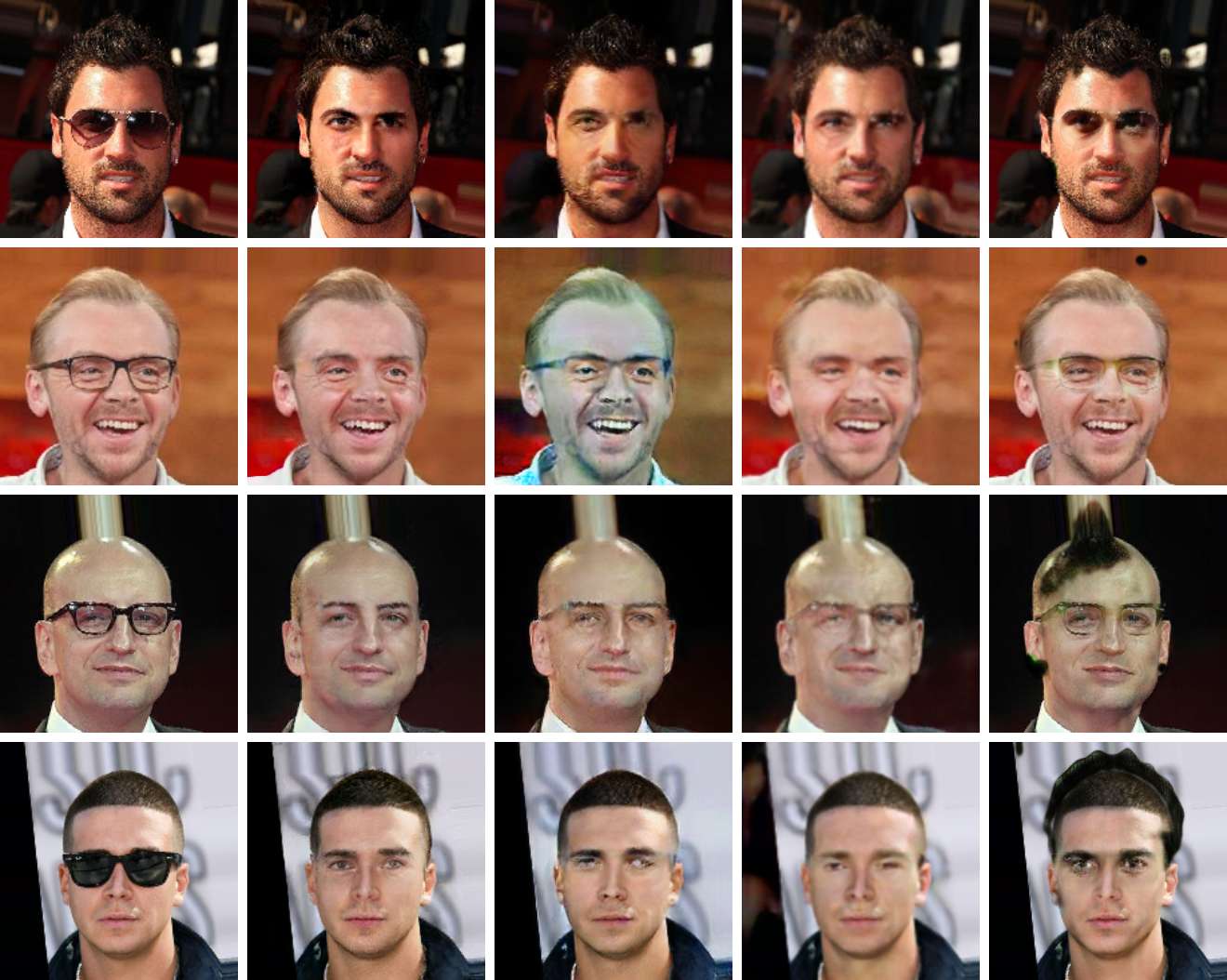}
\caption{Generated images of SPatchGAN and the baselines for glasses removal.}
\label{fig:cmp_glasses}
\end{figure}

We compare SPatchGAN to the baselines, and summarize the quantitative results in Table \ref{tab:vs_baselines}. The qualitative results are shown in Figure \ref{fig:cmp_s2a}, Figure \ref{fig:cmp_m2f} and Figure \ref{fig:cmp_glasses}.

\textbf{Selfie-to-Anime.} Selfie to anime is an application that requires significant shape and texture change. The generated images are shown in Figure \ref{fig:cmp_s2a}. Our method generally achieves the desirable shape deformation, \eg, the anime-style bangs, the reduced height of the face, and two enlarged eyes with roughly the same size. In contrast, other methods often cause incomplete hairstyle change, inconsistent sizes or color of the two eyes, and oversimplified texture. Our method outperforms all the baselines in term of FID and KID in Table \ref{tab:vs_baselines}. The improved image structure of our method comes from the global view of the statistical features, and the relaxed constraints on the generator.

\textbf{Male-to-Female.} To translate a male face to a female face, the major challenge is to change the hairstyle. Some minor modification are also needed for other parts such as the skin and the lips. The generated images are shown in Figure \ref{fig:cmp_m2f}. Similar to the selfie-to-anime application, our model does a better job than other models for changing the hairstyle, showcasing the outstanding capability for shape deformation. Our model also manages to remove the beard, smooth the skin and color the lips naturally. The improved local details show the effectiveness of feature matching at the low levels. We achieve much better quantitative results than the baselines according to Table \ref{tab:vs_baselines}.

\textbf{Glasses Removal.} Different from the aforementioned two applications, we expect to change only a small area of the image for glasses removal. The generated images are shown in Figure \ref{fig:cmp_glasses}. Our method generally has less traces of the glasses on the generated images. This is mainly due to the relaxed cycle constraints. Our model is less likely to generate collapsed eyes for the inputs with sunglasses, thanks to the improved stability. Furthermore, we often observe undesirable changes outside the area of glasses with the baselines. Though we have relaxed the cycle constraints, we can still suppress the redundant changes with the weak cycle loss and the identity loss.  Again, our model achieves the best FID and KID in Table \ref{tab:vs_baselines}.

%% file: sections/conclusions.tex
\section{Conclusions}

In this paper, we have proposed an SPatchGAN discriminator for unsupervised image translation. Our discriminator stabilizes the network by statistical feature matching at multiple scales. It also enables training with relaxed constraints. We have shown in the experiments that our method improves the quality of the generated images, especially those with a large shape deformation. Our model outperforms the existing methods on both FID and KID metrics.

\textbf{Acknowledgments.} We would like to thank Kang Chen and Kewei Yang for helpful discussions.

%% file: sections/supl.tex
\section{Supplementary Material}
In this document, we provide the detailed network architecture of SPatchGAN, the data augmentation method, the study about the weak cycle constraint, the experimental results of applying the SPatchGAN discriminator to other image translation frameworks, as well as additional comparison results for SPatchGAN and the baselines.

\input{supl/impl_details}
\input{supl/data_augmentation}
\input{supl/study_cycle}
\input{supl/scycle_smunit}
\input{supl/additional_results}

%% file: supl/impl_details.tex
\subsection{Implementation Details}

We describe the details of our network architecture in this section. A convolutional layer with kernel size $p \times p$, stride $q$ and number of output channels $w$ is denoted as K$p$-S$q$-C$w$. A fully connected layer with number of output channels $w$ is denoted as FC$w$. A $2 \times$ nearest-neighbor upsampling layer is denoted as U2. A residual block with a shortcut branch and a residual branch $b$ is denoted as RES($b$). A block $b$ repeated $z$ times is denoted as $b \times z$.

\textbf{Discriminator.} The discriminator consists of a feature extraction block and four scales. Each scale has a downsampling block, an adaptation block and three MLPs. Spectral normalization (SN) and Leaky-ReLU (LReLU) with a slope of 0.2 are used in the discriminator.
\begin{itemize}
\itemsep0em
  \item Feature extraction block: K4-S2-C256-SN-LReLU, K4-S2-C512-SN-LReLU.
  \item Downsampling block: K4-S2-C1024-SN-LReLU.
  \item Adaptation block: (K1-S1-C1024-SN-LReLU) $\times 2$.
  \item MLP: (FC1024-SN-LReLU) $\times 2$, FC1-SN.
\end{itemize}

Given an input tensor $a \in \mathbb{R}^{H \times W \times C}$ with height $H$, width $W$ and number of channels $C$, the statistical feature of uncorrected standard deviation is calculated as
\begin{equation}
\label{eqn:stddev}
s_k = \sqrt{\frac{1}{HW}\sum_{i=1}^H\sum_{j=1}^W{(a_{i,j,k}-\bar{a_k})^2}},
\end{equation}
where $s_k$ is the $k$-th element of the output feature vector $s\in \mathbb{R}^C$, $\bar{a_k}$ is the average value of the $k$-th input feature map, and $a_{i,j,k}$ is the input element at the $i$-th row, $j$-th column of the $k$-th feature map. 

\textbf{Forward generator.} The forward generator consists of a downsampling module, a residual module, and an upsampling module. We use instance normalization (IN) in the downsampling and residual modules, and use layer normalization (LN) in the upsampling module. ReLU is utilized as the activation function except for the output layer, which uses Tanh.
\begin{itemize}
\itemsep0em
  \item Downsampling module: K3-S2-C128-IN-ReLU, K3-S2-C256-IN-ReLU, K3-S2-C512-IN-ReLU.
  \item Residual module: RES(K3-S1-C512-IN-ReLU, K3-S1-C512-IN) $\times 8$.
  \item Upsampling module: U2, (K3-S1-C512-LN-ReLU) $\times 2$, U2, K3-S1-C256-LN-ReLU, U2, K3-S1-C128-LN-ReLU, K3-S1-C3-Tanh.
\end{itemize}

\textbf{Backward generator.} The backward generator has a pre-mixing module, a residual module, and a post-mixing module. The residual module has the same structure as the forward generator.
\begin{itemize}
\itemsep0em
   \item Pre-mixing module: K3-S1-C512-IN.
   \item Post-mixing module: K3-S1-C512-LN-ReLU, K3-S1-C3-Tanh.
\end{itemize}

%% file: supl/data_augmentation.tex
\subsection{Data Augmentation}
For selfie-to-anime, we adopt the data augmentation method in U-GAT-IT that first resizes the images to $286\times286$, then randomly crops the images to $256\times256$. For male-to-female and glasses removal, the images are center cropped to $178\times178$, resized to $256\times256$, and randomly shifted by up to $13$ pixels horizontally and vertically.

We apply color jittering with random brightness offset in $[-0.125, 0.125]$, random hue offset in $[-0.02, 0.02]$, random saturation factor in $[0.8, 1.2]$, and random contrast factor in $[0.8, 1.2]$. All images are also randomly flipped horizontally.

%% file: supl/study_cycle.tex
\begin{figure*}
    \centering
    \begin{subfigure}[b]{0.3\linewidth}
        \centering
        \caption*{\footnotesize \hspace{0.3cm} Input \hspace{0.8cm}  w/ Cycle \hspace{0.5cm}  w/o Cycle \hspace{0.2cm}}
        \includegraphics[width=\linewidth]{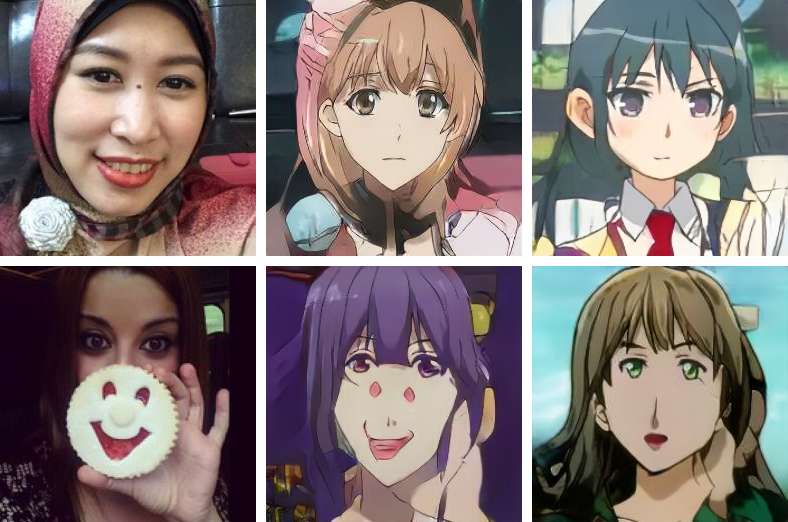}
        \caption{Selfie-to-anime}
        \label{fig:s2a_cyc}
    \end{subfigure}
    \hfill
    \begin{subfigure}[b]{0.3\linewidth}
        \centering
        \caption*{\footnotesize \hspace{0.3cm} Input \hspace{0.8cm}  w/ Cycle \hspace{0.5cm}  w/o Cycle \hspace{0.2cm}}
        \includegraphics[width=\linewidth]{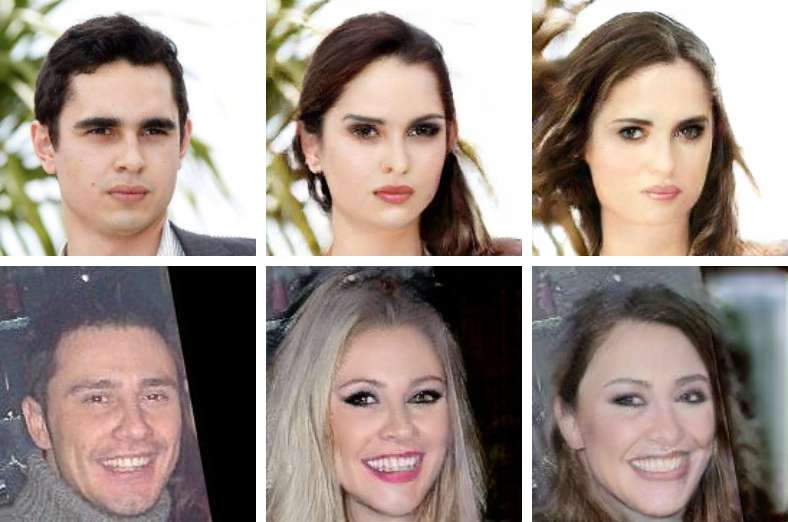}
        \caption{Male-to-female}
        \label{fig:m2f_cyc}
    \end{subfigure}
    \hfill
    \begin{subfigure}[b]{0.3\linewidth}
        \centering
        \caption*{\footnotesize \hspace{0.3cm} Input \hspace{0.8cm}  w/ Cycle \hspace{0.5cm}  w/o Cycle \hspace{0.2cm}}
        \includegraphics[width=\linewidth]{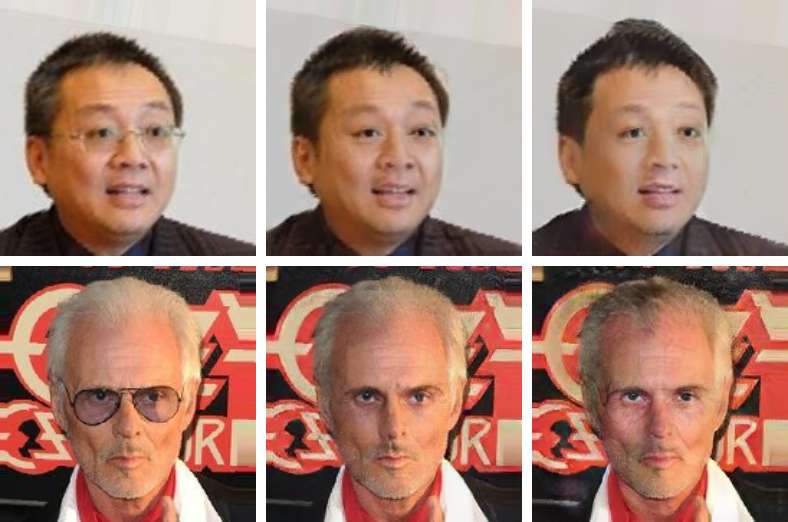}
        \caption{Glasses removal}
        \label{fig:glasses_cyc}
    \end{subfigure}
\caption{Generated images of SPatchGAN with and without the weak cycle constraint. Some redundant changes can be observed after removing the weak cycle constraint.}
\label{fig:weak_cyc}
\end{figure*}

\begin{table*}
\centering
  \begin{tabular}{|c|c|c|c|c|c|c|}
    \hline
    \multirow{2}{*}{Model} &
      \multicolumn{2}{c|}{Selfie-to-Anime} &
      \multicolumn{2}{c|}{Male-to-Female} &
      \multicolumn{2}{c|}{Glasses Removal} \\ \cline{2-7}
    & FID & KID & FID & KID & FID & KID \\
    \hline
    SPatchGAN w/ Weak Cycle & 83.3 & 0.0214 & \textbf{8.73} & \textbf{0.0056} & \textbf{13.9} & \textbf{0.0031} \\
    \hline
    SPatchGAN w/o Weak Cycle & \textbf{82.4} & \textbf{0.0168} & 11.4 & 0.0080 & 16.2 & 0.0047 \\
    \hline
  \end{tabular}
\caption{Quantitative results of SPatchGAN with and without the weak cycle constraint. Lower is better.}
\label{tab:weak_cyc}
\end{table*}

\subsection{Study of the Weak Cycle Constraint}

To further evaluate the stability of SPatchGAN, we try to completely remove the weak cycle constraint. The qualitative results are shown in Figure \ref{fig:weak_cyc}. The generated images still have a good overall quality, verifying that the network has been stabilized to a large extent by the discriminator itself. However, the results without the weak cycle sometimes become too disconnected from the source images. \Eg., the headscarf and the object in front of the face completely disappear in Figure \ref{fig:s2a_cyc}. There are also some undesirable changes of the background in Figure \ref{fig:m2f_cyc}, and some redundant changes of the hair in Figure \ref{fig:glasses_cyc}.

The quantitative results are summarized in Table \ref{tab:weak_cyc}. The FID and KID actually improve for selfie-to-anime after removing the weak cycle. This is partially due to the fact that the similarity between the source and generated images is \emph{not} considered by the metrics. Without the constraint, the images can be translated more freely to match the distributions. We enable the weak cycle in the default setting, since it is desirable to keep the generated image correlated to the source image.

The weak cycle is beneficial for FID and KID in the male-to-female and glasses removal cases. For the applications which aim to adjust only a part of the image, the constraint helps to exclude the unnecessary changes and make the training process more efficient.

Generally speaking, our method helps to separate the need for keeping the source and target images correlated from the need for stabilizing the network. The former is ensured by the weak cycle constraint, while the latter is mainly guaranteed by the SPatchGAN discriminator. Therefore, we can optimize the cycle weight $\lambda^{cyc}$ on an application basis without worrying too much about the stability issues. In contrast, the flexibility of the original cycle based framework is much more limited, since the cycle constraints have to be strict enough to stabilize the network.

%% file: supl/scycle_smunit.tex
\subsection{Applicability to Other Frameworks}

\begin{figure*}
\centering
\caption*{\footnotesize \hspace{0.8cm} Input \hspace{0.9cm}  CycleGAN \hspace{0.5cm}  S-CycleGAN \hspace{0.5cm} MUNIT-1 \hspace{0.7cm} MUNIT-2 \hspace{0.6cm} S-MUNIT-1 \hspace{0.5cm} S-MUNIT-2 \hspace{0.4cm} }
\includegraphics[width=0.8\linewidth]{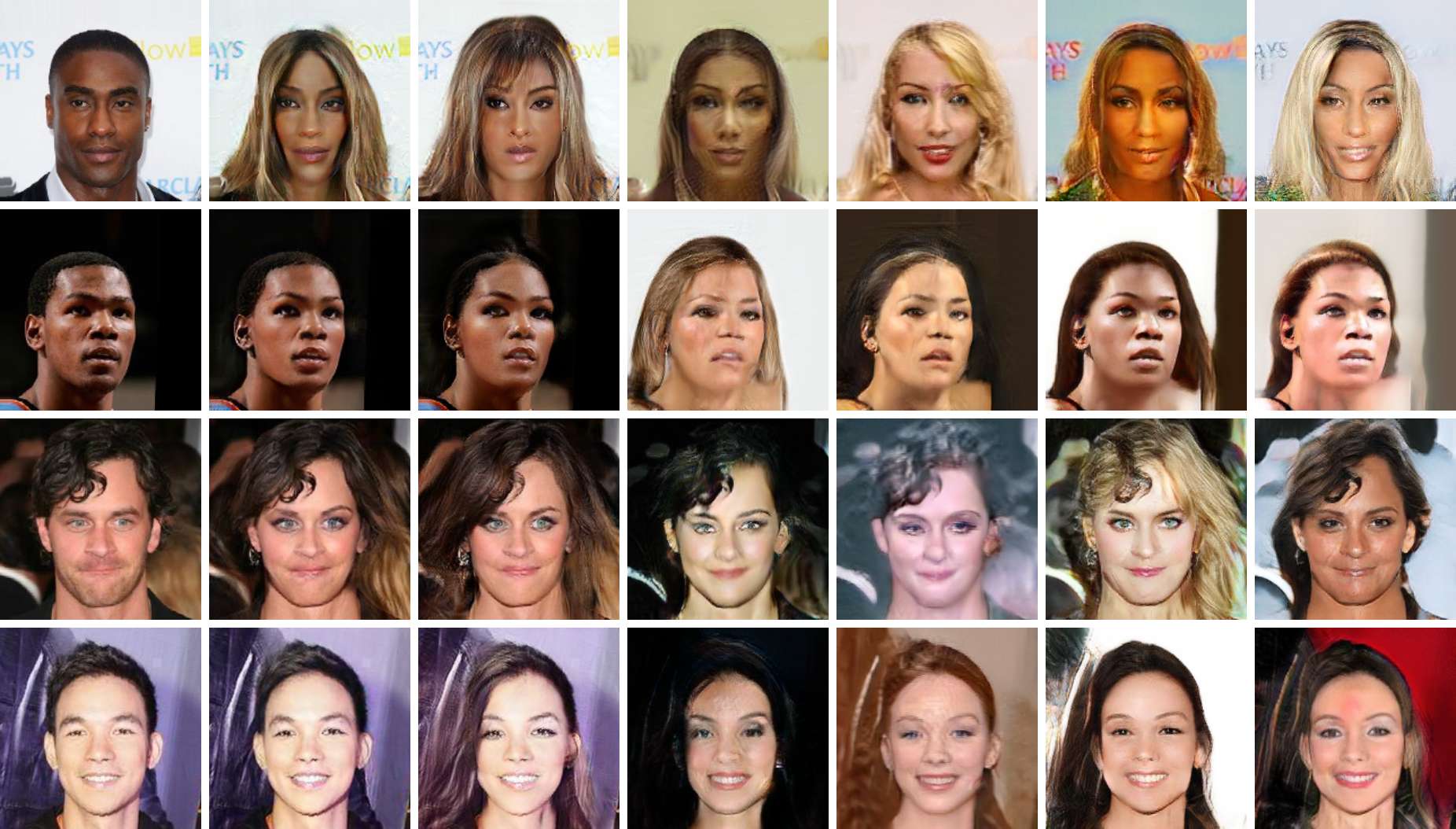}
\caption{Generated images of the applicability studies for male-to-female.}
\label{fig:m2f_scycle_smunit}
\end{figure*}

The SPatchGAN discriminator is generally agnostic of the architecture and constraints for the generator, and can be potentially leveraged to enhance other image translation frameworks. To study its applicability to other frameworks, we directly replace the default discriminators of CycleGAN and MUNIT with the discriminator of SPatchGAN, and evaluate their performance with the male-to-female dataset. The other modules and hyperparameters are unchanged. The qualitative and quantitative results are shown in Figure \ref{fig:m2f_scycle_smunit} and Table \ref{tab:m2f_scycle_smunit}. CycleGAN and MUNIT with the SPatchGAN discriminator are denoted as S-CycleGAN and S-MUNIT respectively. Multimodal results are shown for MUNIT and S-MUNIT.

With the full cycle constraints, the main problem of CycleGAN is the limited shape deformation. In contrast, MUNIT introduces additional stochasticity for multimodality, and suffers more from the instability issue. It can be seen from Figure \ref{fig:m2f_scycle_smunit} that S-CycleGAN helps to make the hairstyle and face more feminine than CycleGAN. S-MUNIT helps to alleviate the blurriness of the generated images compared to MUNIT. The quantitative results of S-CycleGAN and S-MUNIT are also better than CycleGAN and MUNIT according to Table \ref{tab:m2f_scycle_smunit}.

\begin{table}
\begin{center}
\begin{tabular}{|c|c|c|}
\hline
Model                & FID & KID \\ \hline \hline
CycleGAN            & 24.5   & 0.0240  \\ \hline
S-CycleGAN   & 13.4   & 0.0107   \\ \hline
MUNIT    & 20.8   & 0.0161   \\ \hline
S-MUNIT & 17.0   & 0.0123   \\ \hline
\end{tabular}
\caption{Quantitative results of the applicability studies for male-to-female. Lower is better.}
\label{tab:m2f_scycle_smunit}
\end{center}
\end{table}

%% file: supl/additional_results.tex
\subsection{Additional Experimental Results}

\begin{figure*}
\centering
\caption*{\footnotesize \hspace{0.4cm} Input \hspace{0.9cm}  SPatchGAN \hspace{0.4cm}  CycleGAN \hspace{0.7cm} MUNIT \hspace{0.8cm} U-GAT-IT \hspace{1.0cm} CUT \hspace{0.8cm} Council-GAN \hspace{0.3cm} ACL-GAN \hspace{0.1cm} }
\includegraphics[width=0.9\linewidth]{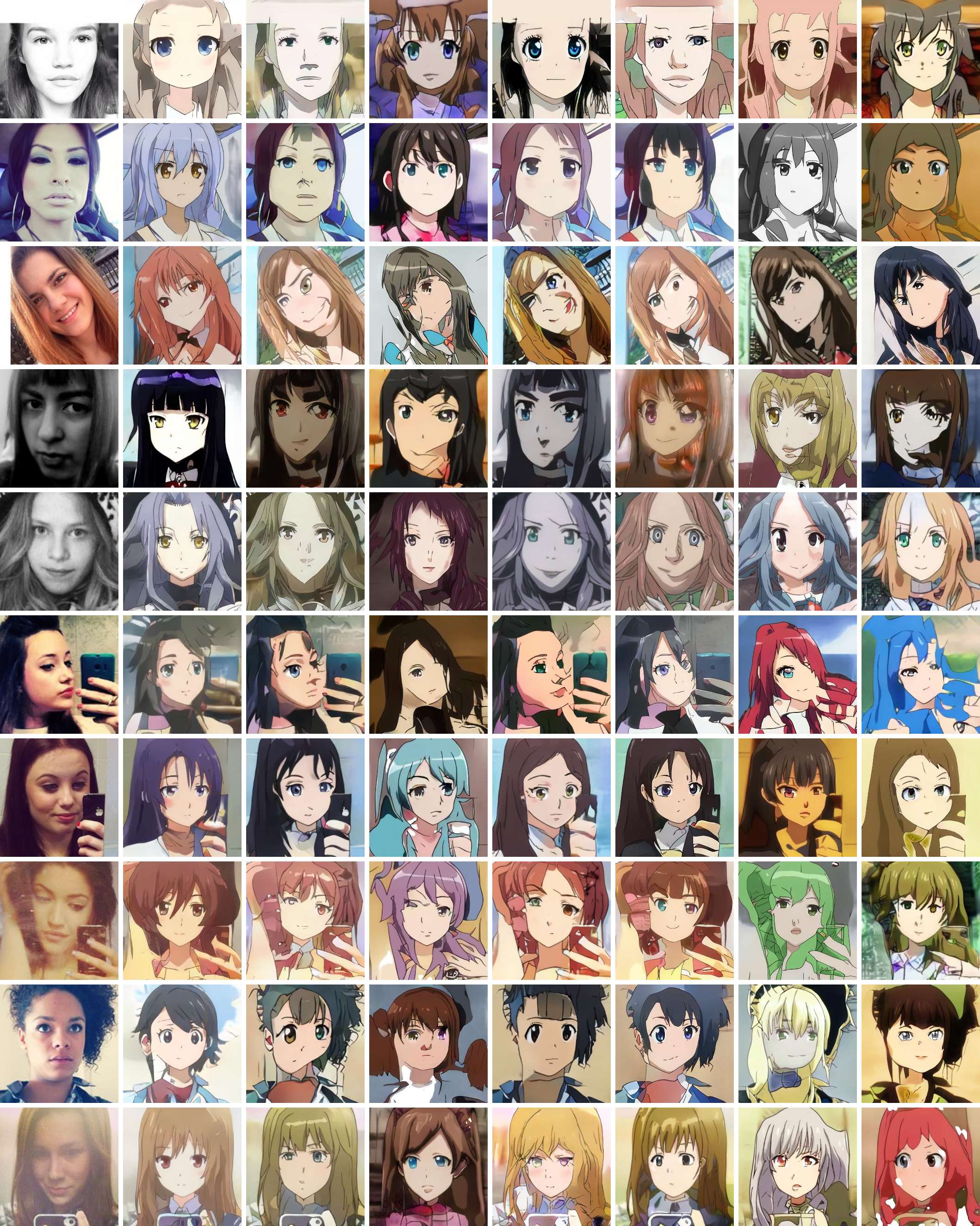}
\caption{Additional results of selfie-to-anime translation.}
\label{fig:supl_s2a}
\end{figure*}

\begin{figure*}
\centering
\caption*{\footnotesize \hspace{0.4cm} Input \hspace{0.9cm}  SPatchGAN \hspace{0.4cm}  CycleGAN \hspace{0.7cm} MUNIT \hspace{0.8cm} U-GAT-IT \hspace{1.0cm} CUT \hspace{0.8cm} Council-GAN \hspace{0.3cm} ACL-GAN \hspace{0.1cm} }
\includegraphics[width=0.9\linewidth]{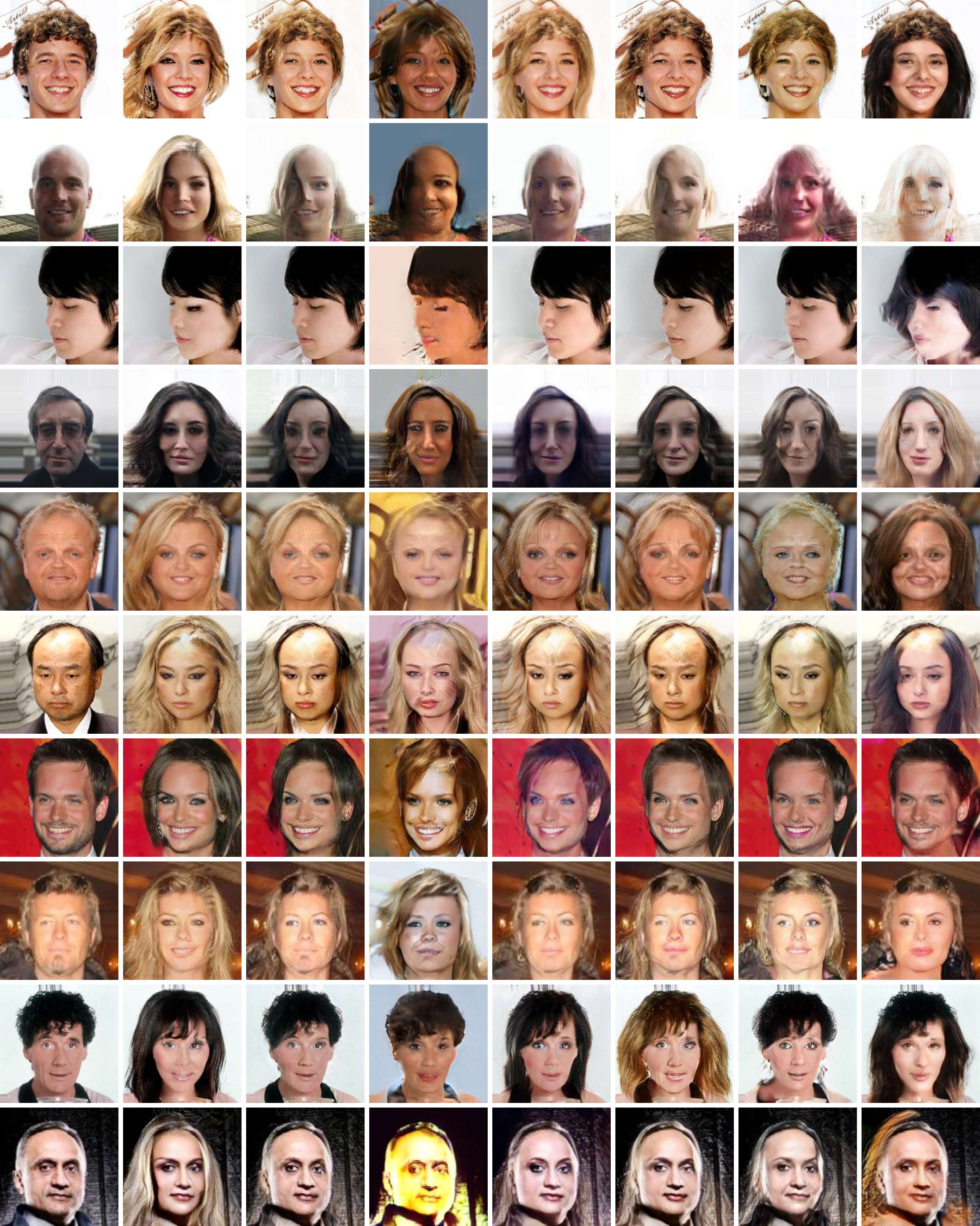}
\caption{Additional results of male-to-female translation.}
\label{fig:supl_m2f}
\end{figure*}

\begin{figure*}
\centering
\caption*{\footnotesize \hspace{0.4cm} Input \hspace{0.9cm}  SPatchGAN \hspace{0.4cm}  CycleGAN \hspace{0.7cm} MUNIT \hspace{0.8cm} U-GAT-IT \hspace{1.0cm} CUT \hspace{0.8cm} Council-GAN \hspace{0.3cm} ACL-GAN \hspace{0.1cm} }
\includegraphics[width=0.9\linewidth]{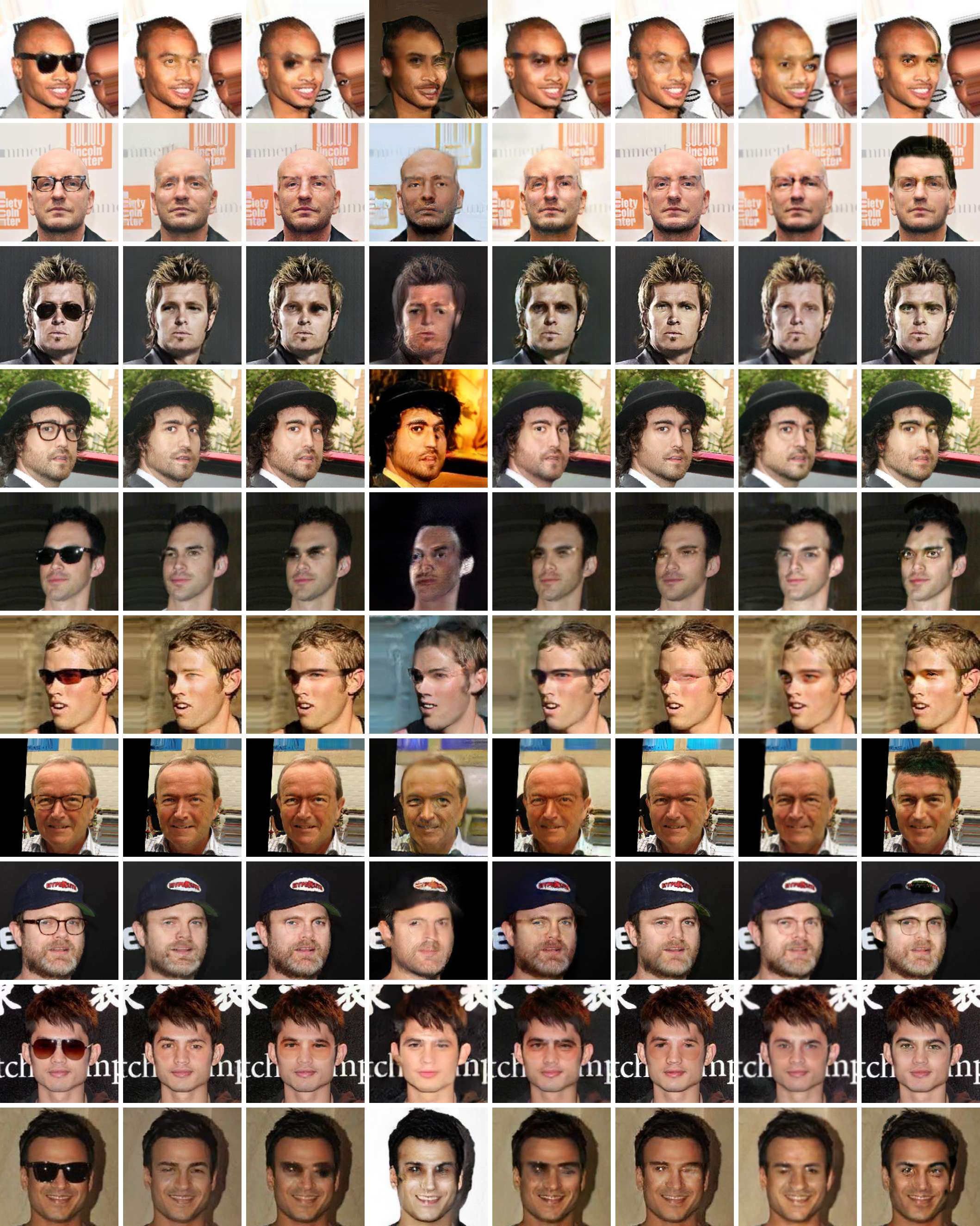}
\caption{Additional results of glasses removal.}
\label{fig:supl_glasses}
\end{figure*}

We show additional results for selfie-to-anime, male-to-female and glasses removal in Figure \ref{fig:supl_s2a}, Figure \ref{fig:supl_m2f} and Figure \ref{fig:supl_glasses}, respectively.

%% file: main.bbl
\begin{thebibliography}{10}\itemsep=-1pt

\bibitem{DBLP:conf/icml/AlmahairiRSBC18}
Amjad Almahairi, Sai Rajeswar, Alessandro Sordoni, Philip Bachman, and Aaron~C.
  Courville.
\newblock Augmented cyclegan: Learning many-to-many mappings from unpaired
  data.
\newblock In {\em Proceedings of the 35th International Conference on Machine
  Learning, {ICML} 2018, Stockholmsm{\"{a}}ssan, Stockholm, Sweden, July 10-15,
  2018}, volume~80, pages 195--204, 2018.

\bibitem{DBLP:journals/corr/ArjovskyCB17}
Mart{\'{\i}}n Arjovsky, Soumith Chintala, and L{\'{e}}on Bottou.
\newblock Wasserstein {GAN}.
\newblock {\em CoRR}, abs/1701.07875, 2017.

\bibitem{DBLP:journals/corr/BaKH16}
Lei~Jimmy Ba, Jamie~Ryan Kiros, and Geoffrey~E. Hinton.
\newblock Layer normalization.
\newblock {\em CoRR}, abs/1607.06450, 2016.

\bibitem{DBLP:conf/iclr/BinkowskiSAG18}
Mikolaj Binkowski, Danica~J. Sutherland, Michael Arbel, and Arthur Gretton.
\newblock Demystifying {MMD} gans.
\newblock In {\em 6th International Conference on Learning Representations,
  {ICLR} 2018, Vancouver, BC, Canada, April 30 - May 3, 2018, Conference Track
  Proceedings}, 2018.

\bibitem{DBLP:conf/iclr/BrockDS19}
Andrew Brock, Jeff Donahue, and Karen Simonyan.
\newblock Large scale {GAN} training for high fidelity natural image synthesis.
\newblock In {\em 7th International Conference on Learning Representations,
  {ICLR} 2019, New Orleans, LA, USA, May 6-9, 2019}, 2019.

\bibitem{DBLP:conf/cvpr/ChenLL18}
Yang Chen, Yu{-}Kun Lai, and Yong{-}Jin Liu.
\newblock Cartoongan: Generative adversarial networks for photo cartoonization.
\newblock In {\em 2018 {IEEE} Conference on Computer Vision and Pattern
  Recognition, {CVPR} 2018, Salt Lake City, UT, USA, June 18-22, 2018}, pages
  9465--9474, 2018.

\bibitem{DBLP:conf/nips/GoodfellowPMXWOCB14}
Ian~J. Goodfellow, Jean Pouget{-}Abadie, Mehdi Mirza, Bing Xu, David
  Warde{-}Farley, Sherjil Ozair, Aaron~C. Courville, and Yoshua Bengio.
\newblock Generative adversarial nets.
\newblock In {\em Advances in Neural Information Processing Systems 27: Annual
  Conference on Neural Information Processing Systems 2014, December 8-13 2014,
  Montreal, Quebec, Canada}, pages 2672--2680, 2014.

\bibitem{DBLP:conf/nips/GulrajaniAADC17}
Ishaan Gulrajani, Faruk Ahmed, Mart{\'{\i}}n Arjovsky, Vincent Dumoulin, and
  Aaron~C. Courville.
\newblock Improved training of wasserstein gans.
\newblock In {\em Advances in Neural Information Processing Systems 30: Annual
  Conference on Neural Information Processing Systems 2017, December 4-9, 2017,
  Long Beach, CA, {USA}}, pages 5767--5777, 2017.

\bibitem{DBLP:conf/nips/HeuselRUNH17}
Martin Heusel, Hubert Ramsauer, Thomas Unterthiner, Bernhard Nessler, and Sepp
  Hochreiter.
\newblock Gans trained by a two time-scale update rule converge to a local nash
  equilibrium.
\newblock In {\em Advances in Neural Information Processing Systems 30: Annual
  Conference on Neural Information Processing Systems 2017, December 4-9, 2017,
  Long Beach, CA, {USA}}, pages 6626--6637, 2017.

\bibitem{DBLP:conf/eccv/HuangLBK18}
Xun Huang, Ming{-}Yu Liu, Serge~J. Belongie, and Jan Kautz.
\newblock Multimodal unsupervised image-to-image translation.
\newblock In {\em Computer Vision - {ECCV} 2018 - 15th European Conference,
  Munich, Germany, September 8-14, 2018, Proceedings, Part {III}}, volume
  11207, pages 179--196, 2018.

\bibitem{DBLP:conf/cvpr/IsolaZZE17}
Phillip Isola, Jun{-}Yan Zhu, Tinghui Zhou, and Alexei~A. Efros.
\newblock Image-to-image translation with conditional adversarial networks.
\newblock In {\em 2017 {IEEE} Conference on Computer Vision and Pattern
  Recognition, {CVPR} 2017, Honolulu, HI, USA, July 21-26, 2017}, pages
  5967--5976, 2017.

\bibitem{DBLP:conf/cvpr/KarnewarW20}
Animesh Karnewar and Oliver Wang.
\newblock {MSG-GAN:} multi-scale gradients for generative adversarial networks.
\newblock In {\em 2020 {IEEE/CVF} Conference on Computer Vision and Pattern
  Recognition, {CVPR} 2020, Seattle, WA, USA, June 13-19, 2020}, pages
  7796--7805, 2020.

\bibitem{DBLP:conf/iclr/KarrasALL18}
Tero Karras, Timo Aila, Samuli Laine, and Jaakko Lehtinen.
\newblock Progressive growing of gans for improved quality, stability, and
  variation.
\newblock In {\em 6th International Conference on Learning Representations,
  {ICLR} 2018, Vancouver, BC, Canada, April 30 - May 3, 2018, Conference Track
  Proceedings}, 2018.

\bibitem{DBLP:conf/cvpr/KarrasLA19}
Tero Karras, Samuli Laine, and Timo Aila.
\newblock A style-based generator architecture for generative adversarial
  networks.
\newblock In {\em {IEEE} Conference on Computer Vision and Pattern Recognition,
  {CVPR} 2019, Long Beach, CA, USA, June 16-20, 2019}, pages 4401--4410, 2019.

\bibitem{DBLP:conf/cvpr/KarrasLAHLA20}
Tero Karras, Samuli Laine, Miika Aittala, Janne Hellsten, Jaakko Lehtinen, and
  Timo Aila.
\newblock Analyzing and improving the image quality of stylegan.
\newblock In {\em 2020 {IEEE/CVF} Conference on Computer Vision and Pattern
  Recognition, {CVPR} 2020, Seattle, WA, USA, June 13-19, 2020}, pages
  8107--8116, 2020.

\bibitem{DBLP:conf/iclr/KimKKL20}
Junho Kim, Minjae Kim, Hyeonwoo Kang, and Kwanghee Lee.
\newblock {U-GAT-IT:} unsupervised generative attentional networks with
  adaptive layer-instance normalization for image-to-image translation.
\newblock In {\em 8th International Conference on Learning Representations,
  {ICLR} 2020, Addis Ababa, Ethiopia, April 26-30, 2020}, 2020.

\bibitem{DBLP:conf/icml/KimCKLK17}
Taeksoo Kim, Moonsu Cha, Hyunsoo Kim, Jung~Kwon Lee, and Jiwon Kim.
\newblock Learning to discover cross-domain relations with generative
  adversarial networks.
\newblock In {\em Proceedings of the 34th International Conference on Machine
  Learning, {ICML} 2017, Sydney, NSW, Australia, 6-11 August 2017}, volume~70,
  pages 1857--1865, 2017.

\bibitem{DBLP:journals/corr/KingmaB14}
Diederik~P. Kingma and Jimmy Ba.
\newblock Adam: {A} method for stochastic optimization.
\newblock In {\em 3rd International Conference on Learning Representations,
  {ICLR} 2015, San Diego, CA, USA, May 7-9, 2015, Conference Track
  Proceedings}, 2015.

\bibitem{DBLP:journals/ijcv/LeeTMHLSY20}
Hsin{-}Ying Lee, Hung{-}Yu Tseng, Qi Mao, Jia{-}Bin Huang, Yu{-}Ding Lu,
  Maneesh Singh, and Ming{-}Hsuan Yang.
\newblock {DRIT++:} diverse image-to-image translation via disentangled
  representations.
\newblock {\em Int. J. Comput. Vis.}, 128(10):2402--2417, 2020.

\bibitem{DBLP:conf/nips/LiuBK17}
Ming{-}Yu Liu, Thomas Breuel, and Jan Kautz.
\newblock Unsupervised image-to-image translation networks.
\newblock In {\em Advances in Neural Information Processing Systems 30: Annual
  Conference on Neural Information Processing Systems 2017, December 4-9, 2017,
  Long Beach, CA, {USA}}, pages 700--708, 2017.

\bibitem{DBLP:conf/iccv/LiuLWT15}
Ziwei Liu, Ping Luo, Xiaogang Wang, and Xiaoou Tang.
\newblock Deep learning face attributes in the wild.
\newblock In {\em 2015 {IEEE} International Conference on Computer Vision,
  {ICCV} 2015, Santiago, Chile, December 7-13, 2015}, pages 3730--3738, 2015.

\bibitem{DBLP:conf/iccv/MaoLXLWS17}
Xudong Mao, Qing Li, Haoran Xie, Raymond Y.~K. Lau, Zhen Wang, and Stephen~Paul
  Smolley.
\newblock Least squares generative adversarial networks.
\newblock In {\em {IEEE} International Conference on Computer Vision, {ICCV}
  2017, Venice, Italy, October 22-29, 2017}, pages 2813--2821, 2017.

\bibitem{DBLP:conf/iclr/MetzPPS17}
Luke Metz, Ben Poole, David Pfau, and Jascha Sohl{-}Dickstein.
\newblock Unrolled generative adversarial networks.
\newblock In {\em 5th International Conference on Learning Representations,
  {ICLR} 2017, Toulon, France, April 24-26, 2017, Conference Track
  Proceedings}, 2017.

\bibitem{DBLP:journals/corr/MirzaO14}
Mehdi Mirza and Simon Osindero.
\newblock Conditional generative adversarial nets.
\newblock {\em CoRR}, abs/1411.1784, 2014.

\bibitem{DBLP:conf/iclr/MiyatoKKY18}
Takeru Miyato, Toshiki Kataoka, Masanori Koyama, and Yuichi Yoshida.
\newblock Spectral normalization for generative adversarial networks.
\newblock In {\em 6th International Conference on Learning Representations,
  {ICLR} 2018, Vancouver, BC, Canada, April 30 - May 3, 2018, Conference Track
  Proceedings}, 2018.

\bibitem{DBLP:conf/cvpr/NizanT20}
Ori Nizan and Ayellet Tal.
\newblock Breaking the cycle - colleagues are all you need.
\newblock In {\em 2020 {IEEE/CVF} Conference on Computer Vision and Pattern
  Recognition, {CVPR} 2020, Seattle, WA, USA, June 13-19, 2020}, pages
  7857--7866, 2020.

\bibitem{DBLP:conf/eccv/ParkEZZ20}
Taesung Park, Alexei~A. Efros, Richard Zhang, and Jun{-}Yan Zhu.
\newblock Contrastive learning for unpaired image-to-image translation.
\newblock In {\em Computer Vision - {ECCV} 2020 - 16th European Conference,
  Glasgow, UK, August 23-28, 2020, Proceedings, Part {IX}}, volume 12354, pages
  319--345, 2020.

\bibitem{DBLP:journals/corr/RadfordMC15}
Alec Radford, Luke Metz, and Soumith Chintala.
\newblock Unsupervised representation learning with deep convolutional
  generative adversarial networks.
\newblock In {\em 4th International Conference on Learning Representations,
  {ICLR} 2016, San Juan, Puerto Rico, May 2-4, 2016, Conference Track
  Proceedings}, 2016.

\bibitem{DBLP:conf/nips/SalimansGZCRCC16}
Tim Salimans, Ian~J. Goodfellow, Wojciech Zaremba, Vicki Cheung, Alec Radford,
  and Xi Chen.
\newblock Improved techniques for training gans.
\newblock In {\em Advances in Neural Information Processing Systems 29: Annual
  Conference on Neural Information Processing Systems 2016, December 5-10,
  2016, Barcelona, Spain}, pages 2226--2234, 2016.

\bibitem{DBLP:conf/iccv/SiddiqueeZTFGBL19}
Md~Mahfuzur~Rahman Siddiquee, Zongwei Zhou, Nima Tajbakhsh, Ruibin Feng,
  Michael~B. Gotway, Yoshua Bengio, and Jianming Liang.
\newblock Learning fixed points in generative adversarial networks: From
  image-to-image translation to disease detection and localization.
\newblock In {\em 2019 {IEEE/CVF} International Conference on Computer Vision,
  {ICCV} 2019, Seoul, Korea (South), October 27 - November 2, 2019}, pages
  191--200, 2019.

\bibitem{DBLP:conf/iclr/TaigmanPW17}
Yaniv Taigman, Adam Polyak, and Lior Wolf.
\newblock Unsupervised cross-domain image generation.
\newblock In {\em 5th International Conference on Learning Representations,
  {ICLR} 2017, Toulon, France, April 24-26, 2017, Conference Track
  Proceedings}, 2017.

\bibitem{DBLP:conf/cvpr/UlyanovVL17}
Dmitry Ulyanov, Andrea Vedaldi, and Victor~S. Lempitsky.
\newblock Improved texture networks: Maximizing quality and diversity in
  feed-forward stylization and texture synthesis.
\newblock In {\em 2017 {IEEE} Conference on Computer Vision and Pattern
  Recognition, {CVPR} 2017, Honolulu, HI, USA, July 21-26, 2017}, pages
  4105--4113, 2017.

\bibitem{DBLP:conf/cvpr/Wang0ZTKC18}
Ting{-}Chun Wang, Ming{-}Yu Liu, Jun{-}Yan Zhu, Andrew Tao, Jan Kautz, and
  Bryan Catanzaro.
\newblock High-resolution image synthesis and semantic manipulation with
  conditional gans.
\newblock In {\em 2018 {IEEE} Conference on Computer Vision and Pattern
  Recognition, {CVPR} 2018, Salt Lake City, UT, USA, June 18-22, 2018}, pages
  8798--8807, 2018.

\bibitem{DBLP:conf/icml/ZhangGMO19}
Han Zhang, Ian~J. Goodfellow, Dimitris~N. Metaxas, and Augustus Odena.
\newblock Self-attention generative adversarial networks.
\newblock In {\em Proceedings of the 36th International Conference on Machine
  Learning, {ICML} 2019, 9-15 June 2019, Long Beach, California, {USA}},
  volume~97, pages 7354--7363, 2019.

\bibitem{DBLP:conf/eccv/ZhaoWD20}
Yihao Zhao, Ruihai Wu, and Hao Dong.
\newblock Unpaired image-to-image translation using adversarial consistency
  loss.
\newblock In {\em Computer Vision - {ECCV} 2020 - 16th European Conference,
  Glasgow, UK, August 23-28, 2020, Proceedings, Part {IX}}, volume 12354, pages
  800--815, 2020.

\bibitem{DBLP:conf/cvpr/ZhouKLOT16}
Bolei Zhou, Aditya Khosla, {\`{A}}gata Lapedriza, Aude Oliva, and Antonio
  Torralba.
\newblock Learning deep features for discriminative localization.
\newblock In {\em 2016 {IEEE} Conference on Computer Vision and Pattern
  Recognition, {CVPR} 2016, Las Vegas, NV, USA, June 27-30, 2016}, pages
  2921--2929, 2016.

\bibitem{DBLP:conf/iccv/ZhuPIE17}
Jun{-}Yan Zhu, Taesung Park, Phillip Isola, and Alexei~A. Efros.
\newblock Unpaired image-to-image translation using cycle-consistent
  adversarial networks.
\newblock In {\em {IEEE} International Conference on Computer Vision, {ICCV}
  2017, Venice, Italy, October 22-29, 2017}, pages 2242--2251, 2017.

\end{thebibliography}
